\documentclass[a4paper,fleqn]{cas-dc}

\usepackage[numbers,sort&compress]{natbib}
\usepackage{amssymb}
\usepackage{amsmath}
\usepackage{ragged2e}
\usepackage{hyperref}
\usepackage{multirow}
\usepackage{threeparttable}
\usepackage{pdflscape}
\usepackage{enumitem}
\usepackage{afterpage}

\def\tsc#1{\csdef{#1}{\textsc{\lowercase{#1}}\xspace}}
\tsc{WGM}
\tsc{QE}
\tsc{EP}
\tsc{PMS}
\tsc{BEC}
\tsc{DE}

\begin{document}
\let\WriteBookmarks\relax
\def\floatpagepagefraction{1}
\def\textpagefraction{.001}
\shorttitle{MMME: Multi-Modal Micro-Expression Dataset}
\shortauthors{C Ma et~al.}

\title [mode = title]{MMME: A Spontaneous Multi-Modal Micro-Expression Dataset Enabling Visual-Physiological Fusion}

\author[1]{Chuang Ma}[orcid=0000-0002-5741-2036]
\ead{ustb_machuang@163.com}

\credit{Conceptualization of this study, Methodology, Writing - Original draft preparation}

\affiliation[1]{organization={The Defense Innovation Institute, Academy of Military Sciences}, city={Beijing}, country={China}}

\author[1]{Yu Pei}
\ead{peiyu1995@buaa.edu.cn}

\credit{Data curation, Formal analysis}

\author[1]{Jianhang Zhang}
\ead{15225746916@163.com}

\credit{Data curation, Investigation}

\author[1]{Shaokai Zhao}
\ead{lnkzsk@126.com}

\credit{Conceptualization of this study, Methodology}

\author[2]{Bowen Ji}
\ead{bwji@nwpu.edu.cn}

\credit{Methodology, Writing - review \& editing}

\affiliation[2]{organization={Northwestern Polytechnical University}, city={Xi'an}, country={China}}

\author[1]{Liang Xie}
\ead{xielnudt@gmail.com}

\credit{Methodology, Validation}

\author[1]{Ye Yan}
\ead{yy_taiic@163.com}

\credit{Conceptualization of this study, Methodology}

\author[1]{Erwei Yin}
\cormark[1]
\ead{yinerwei1985@gmail.com}

\credit{Writing - review \& editing, Validation}



\cortext[cor1]{Corresponding author}

\begin{abstract}
Micro-expressions (MEs) are subtle, fleeting nonverbal cues that reveal an individual’s genuine emotional state. Their analysis has attracted considerable interest due to its promising applications in fields such as healthcare, criminal investigation, and human-computer interaction. However, existing ME research is limited to single visual modality, overlooking the rich emotional information conveyed by other physiological modalities, resulting in ME recognition and spotting performance far below practical application needs. Therefore, exploring the cross-modal association mechanism between ME visual features and physiological signals (PS), and developing a multimodal fusion framework, represents a pivotal step toward advancing ME analysis. This study introduces a novel ME dataset, MMME, which, for the first time, enables synchronized collection of facial action signals (MEs), central nervous system signals (EEG), and peripheral PS (PPG, RSP, SKT, EDA, and ECG). By overcoming the constraints of existing ME corpora, MMME comprises 634 MEs, 2,841 macro-expressions (MaEs), and 2,890 trials of synchronized multimodal PS, establishing a robust foundation for investigating ME neural mechanisms and conducting multimodal fusion-based analyses. Extensive experiments validate the dataset’s reliability and provide benchmarks for ME analysis, demonstrating that integrating MEs with PS significantly enhances recognition and spotting performance. To the best of our knowledge, MMME is the most comprehensive ME dataset to date in terms of modality diversity. It provides critical data support for exploring the neural mechanisms of MEs and uncovering the visual-physiological synergistic effects, driving a paradigm shift in ME research from single-modality visual analysis to multimodal fusion. The dataset will be publicly available upon acceptance of this paper.
\end{abstract}


\maketitle

\section{Introduction}
\label{sec:introduction}

Emotions play a pivotal role in regulating human behavior and cognition, significantly influencing decision-making, social interactions, and mental health \cite{pessoa2008relationship}. As a complex and multimodal phenomenon, emotions are expressed through a variety of external and internal activities \cite{ezzameli2023emotion}. External expressions encompass facial expressions, verbal communication, and body language, while internal responses include physiological changes such as central nervous system signals (\textit{e.g.}, electroencephalography) and peripheral nervous system signals (\textit{e.g.}, electrocardiography, electrodermal activity, and respiration). Emotion recognition involves analyzing these data to identify and interpret an individual’s emotional state, with potential applications spanning diverse fields such as human-computer interaction, virtual reality, healthcare, and education.

    \begin{figure}[t]
        \centering
        \includegraphics[width=0.9\linewidth]{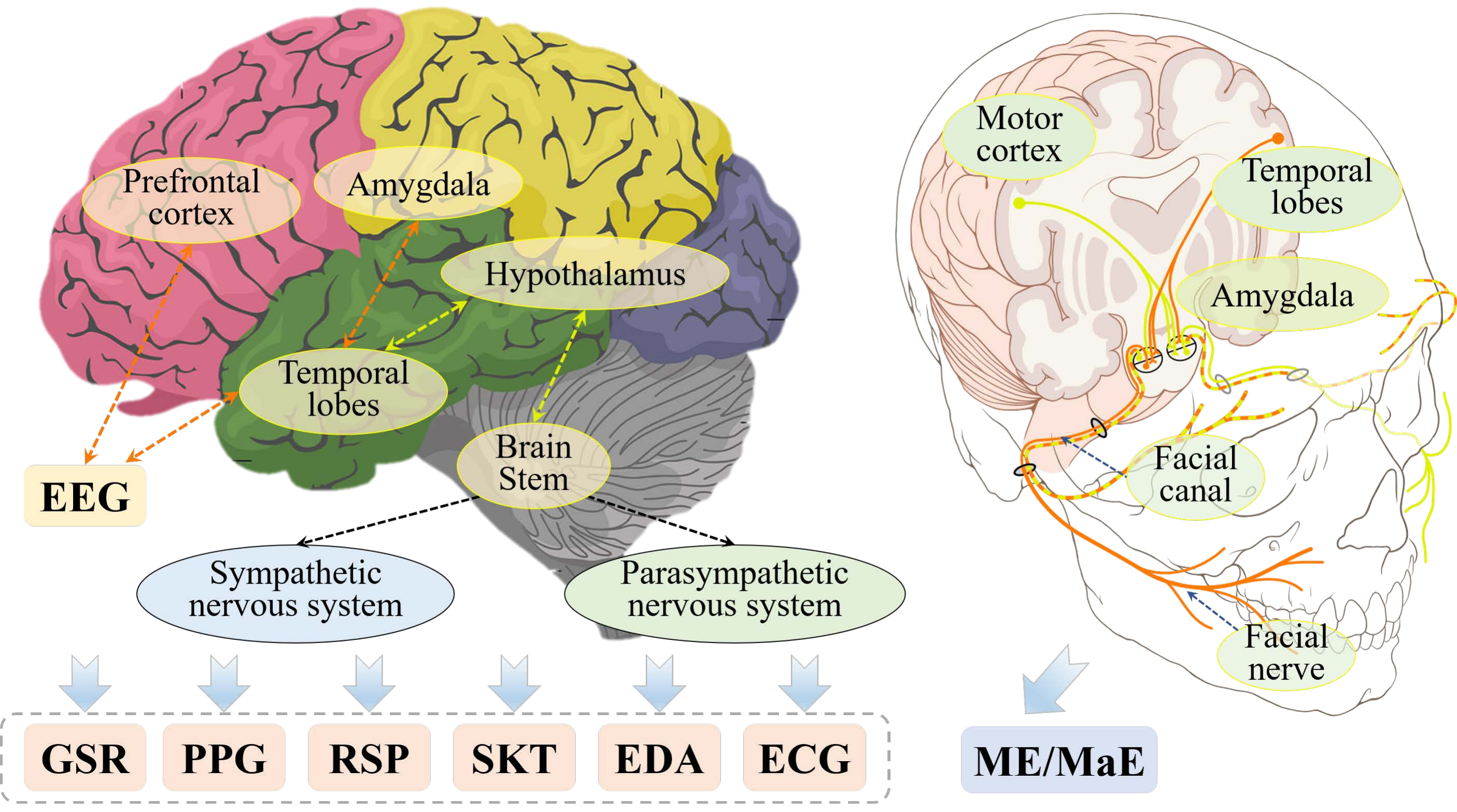}
        \caption{Signal generation and transmission pathway of emotional responses: from central nervous system activation (\textit{e.g.}, cerebral cortex and amygdala) to autonomic and peripheral nervous system responses, leading to physiological changes and facial micro-expressions.}
        \label{fig:Neural}
    \end{figure}

In contemporary society, cultural norms, social expectations, and professional demands often compel individuals to conceal or suppress their genuine emotions through controlled facial expressions, body language, or verbal responses, making it challenging to accurately identify an individual's emotional states \cite{luoma2022interpersonal}. According to Ekman's theory of emotion \cite{ekman1969nonverbal}, suppressed emotions may be unconsciously revealed through brief and subtle facial muscle movements, known as micro-expressions (MEs). MEs are fleeting facial changes, lasting only 1/25 to 1/2 a second (s), often too subtle to be detected by the naked eye \cite{li2022deep}. These involuntary expressions typically surface when individuals attempt to conceal their true emotions, especially in high-pressure scenarios such as interrogations, negotiations, or intense social interactions. From a neurobiological perspective, MEs result from the interplay between voluntary facial expressions, controlled by the pyramidal system, and involuntary expressions, governed by the extrapyramidal system \cite{rinn1984neuropsychology, lee2020neuroanatomy}, as illustrated in Fig. \ref{fig:Neural}. This distinctive neurophysiological mechanism positions MEs as a vital indicator of psychological and physiological stress responses \cite{xie2022overview}. Concurrently, when individuals experience emotional fluctuations, their physiological signals (PS) also exhibit significant patterns of change \cite{kumar2024deep}. For instance, electroencephalography (EEG) records brain activity through scalp electrodes, capturing dynamic changes in brain activity across different emotional states \cite{samal2024role, houssein2022human, liu2023brain}. Electrocardiography (ECG) measures the heart's electrical activity via skin electrodes, reflecting heart rate variability (HRV) and arrhythmias that are closely linked to emotional states \cite{nita2022new}. Additionally, electrodermal activity (EDA), which indirectly measures sweat gland activity, serves as a key indicator of sympathetic nervous system activation and is widely used in deception detection \cite{mercado2024eda}. Respiration (RSP) patterns correlate with emotions: irregular breathing often signals negative arousal (\textit{e.g.}, tension or fear), while slower rates suggest relaxation \cite{chang2024pupil}. Skin temperature (SKT) is a useful predictor of emotional states mediated by the autonomic nervous system and is frequently associated with heightened tension or excitement \cite{saganowski2022emotion}.
Li \textit{et al.} \cite{li2022cas} argued that the high ecological validity of multimodal data facilitated further research on the implementation of ME analysis in practical applications and physiologically-based ME mechanisms. Similarly, Zhang \textit{et al.} \cite{zhang2022short} suggested that the integration of multimodal signals in ME analysis has the potential to overcome certain limitations inherent in existing visual corpora. From a multimodal fusion perspective, combining MEs with PS, including both central and peripheral nervous system signals, not only enables a more comprehensive understanding of the relationships between MEs and PS but also improves the accuracy of emotion recognition. Additionally, it increases the robustness of systems in real-world scenarios where data may be incomplete or ambiguous. 

However, in the field of ME analysis, despite the increasing emphasis on multimodal fusion in recent years, research on MEs and PS continues to proceed largely in isolation, lacking a systematic framework for integrated analysis. This research gap can be attributed to two key factors. First, there is currently a lack of publicly available comprehensive ME datasets that incorporate synchronously acquired MEs and multiple PS. Second, the intrinsic relationship and concordance between MEs and PS have yet to be fully explored and validated. Therefore, developing a spontaneous multimodal ME dataset that synchronously records facial MEs and multiple PS would provide a crucial data foundation for investigating their underlying relationships and neurophysiological mechanisms, offering both theoretical insights and practical applications.
Constructing a high-quality multimodal emotion dataset with scientific value requires careful consideration of key factors, including data synchronization, ecological validity, and sample size. First, ensuring synchronized data collection is essential. The variations in MEs and PS are often transient and rapid. MEs typically last between 40 milliseconds (ms) and 500 ms, while physiological signals, such as EEG and ECG, are recorded at high sampling rates, often reaching the kilohertz range. Therefore, synchronized acquisition ensures precise temporal alignment between different modalities, facilitating the analysis of their dynamic interrelationships. Second, the ecological validity of emotion induction experiments plays a crucial role in data quality. The natural elicitation of MEs is particularly challenging, as their occurrence is influenced by multiple factors, including the experimental environment, stimulus materials, and individual differences \cite{zhao2023dfme}. To enhance data authenticity and reliability, scientifically validated emotion induction paradigms—such as video-based stimuli, facial feedback mechanisms, or immersive interactive scenarios—should be employed. Finally, sample size directly impacts dataset generalizability and model robustness. Given the low occurrence rate of MEs, data collection processes are prone to class imbalance issues \cite{zhang2022balance}. Thus, it is essential to construct a dataset that is not only sufficiently large but also well-balanced across categories to ensure the model's stability and applicability. 

Building upon the above considerations, we developed a multimodal dataset that synchronously captures human facial behavioral signals (MEs), central neural signals (EEG), and multiple peripheral physiological signals (PERI-PS)---including ECG, EDA, RSP, SKT, and PPG---to explore the advantages of multimodal fusion for emotion recognition. The dataset was validated and evaluated using several state-of-the-art algorithms. Additionally, we performed a comprehensive analysis of the correlations between MEs and PS under emotion-elicited conditions, aiming to elucidate the underlying interaction mechanisms across modalities.

The main contributions of this study are as follows:

        \begin{enumerate}
        \item Development of the MMME Dataset: We constructed a multimodal ME dataset, referred to as MMME, which integrates facial MEs, EEG, and multiple PERI-PS. This dataset not only overcame the limitations of relying solely on visual signals for ME analysis but also facilitated the development of more comprehensive and accurate emotion representation models. Furthermore, it enabled an exploration of the dynamic fusion mechanisms underlying emotional changes. 
        
        \item Exploring the concordance between MEs and PS: To effectively integrate visual and physiological features, we investigated whether MEs and PS exhibited synchronized variations during emotional responses. This analysis provided insights into their intrinsic relationships and underlying mechanisms, validating the applicability of the multichannel expression of emotion theory in the field of MEs.

        \item Validation of dataset effectiveness: We conducted ME recognition and spotting tasks on the proposed dataset, employing both unimodal and multimodal fusion approaches. The experimental results demonstrate the efficacy of our multimodal fusion strategy in ME analysis, confirming the dataset's high quality and establishing a reliable benchmark. 
        \end{enumerate}

\section{Related Work}

\subsection{ME Databases}

\begin{table*}[b]
    \centering
    \Large
    \caption{The current spontaneous micro-expression databases. The information about our dataset is provided in the last row}
    \label{tab:ME-databases}
    \renewcommand{\arraystretch}{1.2} 
    \resizebox{\textwidth}{!}{%
        \begin{tabular}{|c|c|cc|cccc|c|cc|l|}
            \hline
            \multirow{2}{*}{\textbf{Database}} & \multirow{2}{*}{\textbf{Year}} & \multicolumn{2}{c|}{\textbf{Subjects}} & \multicolumn{4}{c|}{\textbf{Characteristics of the Sample}} & \multirow{2}{*}{\textbf{Equipment}} & \multicolumn{2}{c|}{\textbf{Annotation}} & \multicolumn{1}{c|}{\multirow{2}{*}{\textbf{Download URL}}} \\ \cline{3-8} \cline{10-11}
             &  & \multicolumn{1}{c|}{\textbf{Num}} & \textbf{Eth} & \multicolumn{1}{c|}{\textbf{Num}} & \multicolumn{1}{c|}{\textbf{Resolution}} & \multicolumn{1}{c|}{\textbf{Sampling rate}} & \textbf{Modality} &  & \multicolumn{1}{c|}{\textbf{Emotion}} & \textbf{AU} & \multicolumn{1}{c|}{} \\ \hline
            \multirow{3}{*}{SMIC \cite{pfister2011recognising}} & \multirow{3}{*}{2013} & \multicolumn{1}{c|}{16} & \multirow{3}{*}{Y} & \multicolumn{1}{c|}{164} & \multicolumn{1}{c|}{$640 \times 480$} & \multicolumn{1}{c|}{100 fps} & HS-RGB & \begin{tabular}[c]{@{}c@{}}PixeLINK\\ L-B774U camera\end{tabular} & \multicolumn{1}{c|}{Pos(51) Neg(70) Sur(43)} & \multirow{3}{*}{-} & \multirow{3}{*}{\begin{tabular}[c]{@{}l@{}}https://www.oulu.fi/en/university/faculties-and-units \\ /faculty-information-technology-and-electrical-engineering \\ /center-for-machine-vision-and-signal-analysis \end{tabular}} \\ \cline{3-3} \cline{5-10}  &  & \multicolumn{1}{c|}{8} &  & \multicolumn{1}{c|}{71} & \multicolumn{1}{c|}{$640 \times 480$} & \multicolumn{1}{c|}{25 fps} & RGB & Standard web camera & \multicolumn{1}{c|}{Pos(28) Neg(23) Sur(20)} &  &  \\ \cline{3-3} \cline{5-10}  &  & \multicolumn{1}{c|}{8} &  & \multicolumn{1}{c|}{71} & \multicolumn{1}{c|}{$640 \times 480$} & \multicolumn{1}{c|}{25 fps} & NIR & Near-infrared camera & \multicolumn{1}{c|}{Pos(28) Neg(24) Sur(20)} &  &  \\ \hline
            \multirow{3}{*}{CASME \cite{yan2013casme}} & \multirow{3}{*}{2013} & \multicolumn{1}{c|}{\multirow{3}{*}{19}} & \multirow{3}{*}{N} & \multicolumn{1}{c|}{\multirow{3}{*}{195}} & \multicolumn{1}{c|}{$1,280 \times 720$} & \multicolumn{1}{c|}{60 fps} & RGB & BenQ M31 camera & \multicolumn{1}{c|}{\multirow{2}{*}{\begin{tabular}[c]{@{}c@{}}Amu(5) Dis(88) Fea(2)\\ Con(3) Sad(6) Ten(28)\\ Sur(20) Rep(40)\end{tabular}}} & \multirow{3}{*}{12+} & \multirow{3}{*}{\begin{tabular}[c]{@{}l@{}}http://casme.psych.ac.cn/casme/c1\end{tabular}} \\ \cline{6-9}  &  & \multicolumn{1}{c|}{} &  & \multicolumn{1}{c|}{} & \multicolumn{1}{c|}{$640 \times 480$} & \multicolumn{1}{c|}{60 fps} & RGB & \begin{tabular}[c]{@{}c@{}}Point Grey\\ GRAS-03K2C camera\end{tabular} & \multicolumn{1}{c|}{} &  &  \\ \hline
            CASME II \cite{yan2014casme} & 2014 & \multicolumn{1}{c|}{26} & N & \multicolumn{1}{c|}{247} & \multicolumn{1}{c|}{$640 \times 480$} & \multicolumn{1}{c|}{200 fps} & RGB & \begin{tabular}[c]{@{}c@{}}Point Grey\\ GRAS-03K2C camera\end{tabular} & \multicolumn{1}{c|}{\begin{tabular}[c]{@{}c@{}}Hap(33) Dis(60) Sur(25)\\ Rep(27) Oth(102)\end{tabular}} & 11+ & \begin{tabular}[c]{@{}l@{}}http://casme.psych.ac.cn/casme/c2\end{tabular} \\ \hline
            SAMM \cite{davison2016samm} & 2016 & \multicolumn{1}{c|}{32} & Y & \multicolumn{1}{c|}{159} & \multicolumn{1}{c|}{$2,040 \times 1,080$} & \multicolumn{1}{c|}{200 fps} & Grayscale & \begin{tabular}[c]{@{}c@{}}Basler Ace\\ acA2000-340km camera\end{tabular} & \multicolumn{1}{c|}{\begin{tabular}[c]{@{}c@{}}Hap(24) Dis(8) Fea(7)\\ Ang(20) Sur(13) Sad(3)\\ Oth(84)\end{tabular}} & ALL & \begin{tabular}[c]{@{}l@{}}https://helward.mmu.ac.uk\\ /STAFF/M.Yap/dataset.php\end{tabular} \\ \hline
            MEVIEW \cite{husak2017spotting} & 2017 & \multicolumn{1}{c|}{16} & N & \multicolumn{1}{c|}{29} & \multicolumn{1}{c|}{$1,280 \times 720$} & \multicolumn{1}{c|}{30 fps} & RGB & - & \multicolumn{1}{c|}{\begin{tabular}[c]{@{}c@{}}Hap(4) Dis(1) Fea(3)\\ Ang(1) Sur(8) Con(4)\\ Unc(7)\end{tabular}} & 7 & \begin{tabular}[c]{@{}l@{}}https://cmp.felk.cvut.cz/~cechj/ME/\end{tabular} \\ \hline
            CAS(ME)$^{2}$ \cite{qu2017cas} & 2018 & \multicolumn{1}{c|}{22} & N & \multicolumn{1}{c|}{57} & \multicolumn{1}{c|}{$640 \times 480$} & \multicolumn{1}{c|}{30 fps} & RGB & \begin{tabular}[c]{@{}c@{}}Logitech Pro\\ C920 camera\end{tabular} & \multicolumn{1}{c|}{\begin{tabular}[c]{@{}c@{}}Pos(8) Neg(21) Sur(9)\\ Oth(19)\end{tabular}} & 28 & \begin{tabular}[c]{@{}l@{}}http://casme.psych.ac.cn/casme/c3\end{tabular} \\ \hline
            MMEW \cite{ben2021video} & 2021 & \multicolumn{1}{c|}{36} & N & \multicolumn{1}{c|}{300} & \multicolumn{1}{c|}{$1,920 \times 1,080$} & \multicolumn{1}{c|}{90 fps} & RGB & - & \multicolumn{1}{c|}{\begin{tabular}[c]{@{}c@{}}Hap(36) Dis(72) Fea(16)\\ Ang(8) Sur(89) Sad(13)\\ Oth(66)\end{tabular}} & 17 & \begin{tabular}[c]{@{}l@{}}https://github.com/benxianyeteam\\/MMEW-Dataset\end{tabular} \\ \hline
            \multirow{5}{*}{4DME \cite{li20224dme}} & \multirow{5}{*}{2022} & \multicolumn{1}{c|}{\multirow{5}{*}{65}} & \multirow{5}{*}{Y} & \multicolumn{1}{c|}{\multirow{5}{*}{1068}} & \multicolumn{1}{c|}{$1,200 \times 1,600$} & \multicolumn{1}{c|}{60 fps} & 4D & \begin{tabular}[c]{@{}c@{}}Basler \\ avA1600 65k  camera\end{tabular} & \multicolumn{1}{c|}{\multirow{4}{*}{\begin{tabular}[c]{@{}c@{}}Pos(34) Neg(127) Sur(30)\\ Rep(6) PosSur(13) NegSur(8)\\ RepSur(3) PosRep(8)\\ NegRep(7) Oth(31)\end{tabular}}} & \multirow{5}{*}{22} & \multirow{5}{*}{\begin{tabular}[c]{@{}l@{}}Contact: \\ xiaobai.li@oulu.fi\end{tabular}} \\ \cline{6-9}
             &  & \multicolumn{1}{c|}{} &  & \multicolumn{1}{c|}{} & \multicolumn{1}{c|}{$640 \times 480$} & \multicolumn{1}{c|}{60 fps} & Grayscale & \begin{tabular}[c]{@{}c@{}}Stingray\\ F-046B camera\end{tabular} & \multicolumn{1}{c|}{} &  &  \\ \cline{6-9}
             &  & \multicolumn{1}{c|}{} &  & \multicolumn{1}{c|}{} & \multicolumn{1}{c|}{$640 \times 480$} & \multicolumn{1}{c|}{30 fps} & RGB & \multirow{2}{*}{\begin{tabular}[c]{@{}c@{}}Kinect\\ Xbox 360 camera\end{tabular}} & \multicolumn{1}{c|}{} &  &  \\ \cline{6-8}
             &  & \multicolumn{1}{c|}{} &  & \multicolumn{1}{c|}{} & \multicolumn{1}{c|}{$640 \times 480$} & \multicolumn{1}{c|}{30 fps} & Depth &  & \multicolumn{1}{c|}{} &  &  \\ \hline
            \multirow{7}{*}{CAS(ME)$^{3}$ \cite{li2022cas}} & \multirow{7}{*}{2022} & \multicolumn{1}{c|}{\multirow{7}{*}{247}} & \multirow{7}{*}{N} & \multicolumn{1}{c|}{\multirow{2}{*}{1109}} & \multicolumn{1}{c|}{\multirow{2}{*}{$1,280 \times 720$}} & \multicolumn{1}{c|}{\multirow{2}{*}{30 fps}} & RGB & \multirow{2}{*}{\begin{tabular}[c]{@{}c@{}}$\rm Intel^{@}$ $\rm RealSense^{TM}$ \\ D415 camera\end{tabular}} & \multicolumn{1}{c|}{\multirow{7}{*}{\begin{tabular}[c]{@{}c@{}}Part A:\\ Hap(64) Dis(281) Fea(93)\\ Ang(70) Sur(201) Sad(64)\\ Oth(170)\\ Part C:\\ Pos(16) Neg(99) Sur(30)\\ Oth(20)\end{tabular}}} & \multirow{7}{*}{ALLE} & \multirow{7}{*}{\begin{tabular}[c]{@{}l@{}}http://casme.psych.ac.cn/casme/c4\end{tabular}} \\ \cline{8-8}
             &  & \multicolumn{1}{c|}{} &  & \multicolumn{1}{c|}{} & \multicolumn{1}{c|}{} & \multicolumn{1}{c|}{} & Depth &  & \multicolumn{1}{c|}{} &  &  \\ \cline{5-9}
             &  & \multicolumn{1}{c|}{} &  & \multicolumn{1}{c|}{\multirow{5}{*}{-}} & \multicolumn{1}{c|}{\multirow{4}{*}{16 bits}} & \multicolumn{1}{c|}{\multirow{4}{*}{200 Hz}} & PPG & \multirow{4}{*}{BIOPAC MP160} & \multicolumn{1}{c|}{} &  &  \\ \cline{8-8}
             &  & \multicolumn{1}{c|}{} &  & \multicolumn{1}{c|}{} & \multicolumn{1}{c|}{} & \multicolumn{1}{c|}{} & RSP &  & \multicolumn{1}{c|}{} &  &  \\ \cline{8-8}
             &  & \multicolumn{1}{c|}{} &  & \multicolumn{1}{c|}{} & \multicolumn{1}{c|}{} & \multicolumn{1}{c|}{} & EDA &  & \multicolumn{1}{c|}{} &  &  \\ \cline{8-8}
             &  & \multicolumn{1}{c|}{} &  & \multicolumn{1}{c|}{} & \multicolumn{1}{c|}{} & \multicolumn{1}{c|}{} & ECG &  & \multicolumn{1}{c|}{} &  &  \\ \cline{6-9}
             &  & \multicolumn{1}{c|}{} &  & \multicolumn{1}{c|}{} & \multicolumn{1}{c|}{-} & \multicolumn{1}{c|}{48,000 Hz} & Audio & Video recorder & \multicolumn{1}{c|}{} &  &  \\ \hline
            DFME \cite{zhao2023dfme} & 2023 & \multicolumn{1}{c|}{671} & N & \multicolumn{1}{c|}{7,526} & \multicolumn{1}{c|}{$1,024 \times 768$} & \multicolumn{1}{c|}{200/300/500 fps} & RGB & \begin{tabular}[c]{@{}c@{}}Self-developed\\ high-speed camera\end{tabular} & \multicolumn{1}{c|}{\begin{tabular}[c]{@{}c@{}}Hap(992) Dis(2528) Fea(892)\\ Ang(619) Sur(1208) Sad(635)\\ Oth(251)\end{tabular}} & 22 & \begin{tabular}[c]{@{}l@{}}https://mea-lab-421.github.io\end{tabular} \\ \hline
            \multirow{9}{*}{\begin{tabular}[c]{@{}c@{}}MMME\\ (Ours)\end{tabular}} & \multirow{9}{*}{2025} & \multicolumn{1}{c|}{\multirow{9}{*}{75}} & \multirow{9}{*}{N} & \multicolumn{1}{c|}{634} & \multicolumn{1}{c|}{$2,040 \times 1,080$} & \multicolumn{1}{c|}{150 fps} & RGB & \begin{tabular}[c]{@{}c@{}}XIMEA\\ MQ022CG-CM camera\end{tabular} & \multicolumn{1}{c|}{\multirow{9}{*}{\begin{tabular}[c]{@{}c@{}}Hap(92) Dis(163) Fea(124)\\ Ang(50) Sur(105) Sad(55)\\ Con(45)\\ Valence and Arousal\end{tabular}}} & \multirow{9}{*}{21} & \multirow{9}{*}{https://github.com/Mac0504/MMME} \\ \cline{5-9}
             &  & \multicolumn{1}{c|}{} &  & \multicolumn{1}{c|}{2,890 trials} & \multicolumn{1}{c|}{24 bits} & \multicolumn{1}{c|}{1,000 Hz} & EEG & \begin{tabular}[c]{@{}c@{}}Neuroscan SynAmps\\ RT EEG system\end{tabular} & \multicolumn{1}{c|}{} &  &  \\ \cline{5-9}
             &  & \multicolumn{1}{c|}{} &  & \multicolumn{1}{c|}{\multirow{5}{*}{2,890 trials}} & \multicolumn{1}{c|}{\multirow{5}{*}{16 bits}} & \multicolumn{1}{c|}{\multirow{5}{*}{1,000 Hz}} & PPG & \multirow{5}{*}{BIOPAC MP160} & \multicolumn{1}{c|}{} &  &  \\ \cline{8-8}
             &  & \multicolumn{1}{c|}{} &  & \multicolumn{1}{c|}{} & \multicolumn{1}{c|}{} & \multicolumn{1}{c|}{} & RSP &  & \multicolumn{1}{c|}{} &  &  \\ \cline{8-8}
             &  & \multicolumn{1}{c|}{} &  & \multicolumn{1}{c|}{} & \multicolumn{1}{c|}{} & \multicolumn{1}{c|}{} & SKT &  & \multicolumn{1}{c|}{} &  &  \\ \cline{8-8}
             &  & \multicolumn{1}{c|}{} &  & \multicolumn{1}{c|}{} & \multicolumn{1}{c|}{} & \multicolumn{1}{c|}{} & EDA &  & \multicolumn{1}{c|}{} &  &  \\ \cline{8-8}
             &  & \multicolumn{1}{c|}{} &  & \multicolumn{1}{c|}{} & \multicolumn{1}{c|}{} & \multicolumn{1}{c|}{} & ECG &  & \multicolumn{1}{c|}{} &  &  \\ \hline
        \end{tabular}
        }       
        \begin{tablenotes}[
            leftmargin=0pt,    
            itemindent=2pt,    
            labelwidth=15pt,   
            labelsep=3pt       
            ]
            \fontsize{7pt}{8pt}\selectfont 
            \item[1] $^1$ \textbf{Year:} The publication year of the dataset's corresponding paper.
            \item[2] $^2$ \textbf{Num:} The number of participants and micro-expression samples is based on the original paper and the downloaded dataset. Note that not all participants \\ successfully exhibited MEs.
            \item[3] $^3$ \textbf{Eth:} Indicates whether the dataset includes participants from multiple ethnic backgrounds. Y: Yes; N: No.
            \item[4] $^4$ \textbf{Resolution:} For cameras, resolution refers to the number of pixels in captured images, which indicates the clarity or level of detail. For physiological signal acquisition devices, resolution represents the smallest distinguishable unit of signal, typically measured in bits.
            \item[5] $^5$ \textbf{Sampling rate:} The camera’s sampling rate is measured in frames per second (fps), while the sampling rate for physiological signals is measured in Hz.
            \item[6] $^6$ \textbf{Modality:} HS-RGB denotes 2D high-speed video; NIR denotes 2D near-infrared video; Grayscale denotes grayscale video; 4D denotes dynamic 3D video; \\ Depth denotes depth images; PPG represents photoplethysmography; RSP represents respiratory signals; EDA represents electrodermal activity; \\ ECG represents electrocardiograms; and EEG represents electroencephalograms.
            \item[7] $^7$ \textbf{Emotion:} \textit{Pos:} Positive; \textit{Neg:} Negative; \textit{Sur:} Surprise; \textit{Amu:} Amusement; \textit{Hap:} Happiness; \textit{Dis:} Disgust; \textit{Rep:} Repression; \textit{Fea:} Fear; \textit{Ang:} Anger; \textit{Con:} \\ Contempt; \textit{Sad:} Sadness; \textit{Ten:} Tense; \textit{PosSur:} Positively Surprise; \textit{NegSur:} Negatively Surprise; \textit{RepSur:} Repressed Surprise; \textit{PosRep:} Positively Repression; \\ \textit{NegRep:} Negatively Repression; \textit{Oth:} Others; \textit{Unc:} Unclear.
            \item[8] $^8$ \textbf{AU:} ALL: all observed AUs; ALLE: all observed AUs except eye blinking.
            \item[9] $^9$ The '-' in the table indicates that the information is not provided in the original paper.
        \end{tablenotes}   
\end{table*}

The advancement of ME analysis technology relies heavily on the availability of spontaneous ME datasets. In recent years, growing interest in MEs within affective computing, combined with an increasing demand for advanced ME analysis, has driven the meticulous development and public release of several ME datasets. Table \ref{tab:ME-databases} offers a detailed overview of these datasets' key features, including release year, participant numbers, sample characteristics, acquisition equipment, annotation details, and accessible download links. The following analysis assesses these datasets in terms of paradigm design, experimental conditions, data modalities, and sample annotation practices.

\subsubsection{Paradigm Design}

Early ME samples were derived from posed, non-sponta-neous datasets, such as the Polikovsky database (2009) \cite{polikovsky2009facial} and the USF-HD dataset (2011) \cite{shreve2011macro}, where participants were instructed to mimic rapid facial expressions. However, these posed MEs failed to capture participants’ authentic emotional states and diverge from the involuntary nature of MEs observed in natural contexts. Consequently, such databases have seen limited application in contemporary ME research. According to the emotional leakage theory \cite{ekman1968nonverbal}, when individuals suppress their emotions to maintain a neutral facade, brief, involuntary facial muscle movements---known as MEs---may emerge, inadvertently revealing the concealed affective states. Building on this insight, researchers have refined experimental approaches by introducing the neutral face paradigm \cite{ben2021video} to elicit spontaneous MEs. In this method, participants are instructed to inhibit facial movements and preserve a neutral expression while viewing emotionally evocative videos. Owing to its controlled environment and practical feasibility, this paradigm has been widely embraced by spontaneous ME datasets, such as SMIC (2011) \cite{pfister2011recognising}, CASME (2013) \cite{yan2013casme}, CASME II (2014) \cite{yan2014casme}, SAMM (2016) \cite{davison2016samm}, CAS(ME)$^{2}$ (2017) \cite{qu2017cas}, MMEW (2021) \cite{ben2021video}, 4DME (2022) \cite{li20224dme}, and DFME (2023) \cite{zhao2023dfme}. Nonetheless, the laboratory-based neutral paradigm differs from real-world conditions, where MEs typically manifest during interpersonal interactions, and individuals may conceal true emotions not only by maintaining neutrality but also by displaying expressions opposite to their genuine feelings or simply smiling.

Recognizing that social interactions elicit more diverse expressions than solitary settings, the MEVIEW dataset \cite{husak2017spotting} compiled video footage from online sources, such as poker games and television interviews, to capture MEs in genuine interactions. Yet, frequent camera movements and angle switches result in fewer samples with complete facial views in this dataset. To enhance ecological validity, the CAS(ME)$^{3}$ (2022) \cite{li2022cas} was constructed in a controlled experimental environment using high-risk simulated crime scenarios to elicit MEs. Researchers evaluated participants' deceptive behaviors during the interrogation phase. However, MEs elicited under this paradigm remain closely linked to deceptive behavior, and their faint signals were often obscured by non-emotional facial movements inherent to dialogue. Based on the neutral facial paradigm, this study developed a Continuous Monitoring and Real-time Reminder (CMRR) experimental paradigm to enhance participants' motivation for facial expression suppression, as analyzed in Section \ref{Procedure}.

\subsubsection{Experimental Equipments}

In terms of experimental equipment, capturing MEs imposes stringent technical demands on imaging devices. Due to the brief duration and low intensity of MEs, high-frame-rate and high-resolution cameras are essential to accurately record subtle changes in facial muscle activity \cite{borza2017high}. However, early MEs datasets were constrained by the limited performance of their recording equipment, with frame rates typically below 100 frames per second (fps) and resolutions often restricted to 640 $\times$ 480 pixels. Specifically, in the first publicly available spontaneous MEs database, SMIC, the HS subset was recorded using a 100 fps camera, while the VIS subset was captured at a mere 25 fps. Similarly, datasets such as MEVIEW, 4DME, CAS(ME)$^{2}$, and CAS(ME)$^{3}$ employed a sampling rate of 30 fps, while CASME used 60 fps and MMEW utilized 90 fps. Notably, at a frame rate of 30 fps, a complete MEs sample captures a maximum of 15 frames, resulting in a relatively coarse retention of temporal variation characteristics \cite{zhang2024review}. To address these limitations and enhance data quality, subsequent studies have adopted higher-performance industrial-grade cameras. For instance, CASME II employed a Point Grey camera with a frame rate of 200 fps, and the SAMM dataset utilized a 200 fps Basler Ace camera. Particularly noteworthy is the DFME dataset, where the research team developed a custom high-speed camera with a configurable frame rate, achieving a maximum sampling speed of 500 fps. Regarding resolution, the lowest in existing public datasets is SMIC’s 640 $\times$ 480 pixels, while the highest reaches 2,040 $\times$ 1,080 pixels in the SAMM dataset. Although high frame rates and resolutions significantly improve the precision of MEs capture, excessively high parameter settings may introduce challenges such as data redundancy, increased transmission bandwidth demands, and elevated storage and processing burdens. Consequently, researchers must strike a balance between accurate capture and data processing efficiency during experimental design to ensure both the quality of MEs data and the feasibility of experimentation.
In the configuration of lighting environments, stable and appropriate illumination conditions play a critical role in determining the quality of MEs data collection. To mitigate image flickering caused by 50 Hz alternating current, studies have predominantly adopted direct current (DC)-powered LED lighting systems. To ensure uniform light distribution across the participant’s facial region, experiments typically combine LED lights with umbrella reflectors. Furthermore, to minimize background interference in micro-expression recognition (MER), experimental setups often incorporate plain-colored backdrops. This facilitates precise facial region segmentation and feature extraction during subsequent image processing stages. In accordance with the aforementioned experimental environment configuration, this study employed an XIMEA industrial-grade high-speed camera for facial image acquisition, with a capture resolution of 2,040 $\times$ 1,080 pixels and a frame rate of 150 fps. To ensure reliable storage of the substantial volume of data, the experimental system was equipped with a 100TB disk array storage device. The detailed configuration of experimental equipment is presented in Section \ref{Equipment}.

\subsubsection{Data Modalities}

ME analysis is a prominent computer vision task, typically performed using facial images or videos. Li \textit{et al.} \cite{li20224dme} argued that existing ME datasets suffer from a lack of data diversity, and the monotonous data formats restrict the application of current MER methods. Zhao \textit{et al.} \cite{zhao2022differences} proposed that investigating the neural mechanisms underlying MEs could lay the foundation for MER based on PS, thereby helping to overcome the limitations of MER and broader its application scenarios. Li \textit{et al.} \cite{li2022deep} further suggested that utilizing multiple modalities could provide complementary information and enhance classification robustness. For example, different emotional expressions can produce distinct changes in autonomic nervous system activity, such as increased heart rate and decreased skin temperature during fear. Therefore, PS can be employed to integrate complementary information for further improving MER. Recent advancements have seen the inclusion of additional modalities in ME datasets. For instance, the SMIC dataset incorporates both RGB and near-infrared (NIR) images to increase data diversity. Experimental results demonstrate that under low-light conditions, the NIR camera exhibits a clear advantage over conventional cameras. The 4DME dataset features RGB images, grayscale images, depth images, and dynamic 3D videos (referred to as 4D) \cite{sandbach2012static}, providing richer data that helps mitigate the noise issues typical in traditional 2D videos, such as those caused by self-occlusion, head movement, and lighting variations. Experimental findings suggest that 4D data offers potential advantages in MER tasks, with the integration of information from multiple views proving more effective than relying on any single view. The CAS(ME)$^{3}$ dataset includes visual information, PS, and speech signals. Experimental results indicate that depth information, as an additional modality, effectively enhances the feature extraction of ME. However, the recognition performance when incorporating EDA or speech signals did not meet expectations. This may be due to insufficient signal denoising and filtering, as well as the inability of speech signals to adequately capture the characteristics of MEs. 

It is important to note that different modalities often have varying sampling frequencies, making data synchronization a critical step in multimodal fusion research. Due to technical limitations, the SMIC dataset exhibits a time delay of approximately 3–5 s between the starting points of the three cameras, requiring manual synchronization during post-processing. In contrast, the CAS(ME)$^{3}$ dataset employs an Intel$^{@}$ RealSense$\rm ^{TM}$ D415 camera to simultaneously record RGB color images and their corresponding depth information, while physiological signals are collected using a BIOPAC MP160 multi-channel physiological recorder. However, the associated publications on this dataset do not explicitly specify the precise synchronization method between physiological and speech signals and the video data. The 4DME dataset, on the other hand, achieves data synchronization across all cameras using a trigger mechanism integrated into its audio capture system. In this study, a shared trigger between the camera and the BIOPAC system was employed to generate timestamps, ensuring precise synchronization between facial image capture and PS acquisition.

\subsubsection{Sample Annotation}

As a crucial step in the construction of ME datasets, precise and scientifically rigorous sample annotation provides reliable classification labels for ME analysis models. The annotation process typically requires multiple professionally trained Facial Action Coding System (FACS) coders to analyze video frames sequentially, identifying the occurrence and duration of facial action units (AUs). The final emotion labels for MEs are then determined by integrating three key sources of information: AU annotations, the emotional category of the elicitation materials, and participants' self-reported emotions.

In the annotation methods used for ME datasets, two primary labeling systems are commonly adopted: discrete emotion models and continuous emotion models. Regarding discrete annotation methods, several well-known datasets employ different classification standards. Notably, SMIC, CAS(ME)$^{2}$, CAS(ME)$^{3}$, and 4DME utilize a three-class classification scheme, categorizing samples into positive, negative, and surprise. This simplified classification strategy not only reduces annotation complexity but also mitigates the issue of class imbalance. To enhance the accuracy of emotion analysis, datasets such as CASME, SAMM, MEVIEW, MMEW, CAS(ME)$^{3}$, and DFME adopt a more detailed seven-class system based on Ekman’s basic emotion theory, classifying emotions into happiness, disgust, fear, anger, surprise, sadness, and others. In contrast to discrete annotation methods, continuous emotion models represent emotional states as coordinate points in a multi-dimensional space. Specifically, this model constructs a two-dimensional coordinate system comprising valence and arousal, where valence reflects the positive or negative nature of an emotion, while arousal represents variations in energy levels from calmness to excitement. Although the SAMM dataset recorded valence and arousal data during experiments, these continuous dimension annotations were not included in the final annotation files. Recognizing the advantages of continuous emotion models in quantifying emotional intensity and their applicability in continuous emotion estimation tasks, the ME dataset constructed in this study integrates both annotation systems. It provides not only traditional discrete emotion labels but also valence and arousal annotations, thereby offering a more comprehensive and fine-grained representation of emotions.

\subsection{ME Analysis Methods}

\subsubsection{ME Recognition Methods}

MER involves identifying emotions conveyed through brief and subtle facial expressions, playing a vital role in affective computing by revealing genuine human emotions. In recent years, the release of publicly available ME datasets has significantly fueled the advancement of automated ME recognition (MER) techniques based on computer vision.

\textbf{MER based on visual information.} ME feature extraction techniques have evolved from early handcrafted computer vision methods to contemporary deep learning approaches, broadly classified into spatial and temporal strategies. In terms of spatial feature extraction, a common approach involves segmenting regions of interest (ROIs) on the face based on the FACS \cite{ekman1978facial}. This method divides the face into multiple regions corresponding to independent muscle groups, followed by appearance normalization for each region \cite{wang2015micro, liu2015main, zhang2021facial}. Polikovsky et al. \cite{polikovsky2009facial} proposed a gradient-based feature that constructs histograms of gradient projections in local regions to describe spatial characteristics. Similarly, the Local Binary Pattern (LBP) \cite{wang2015efficient} operator robustly captures local appearance features by comparing the relative brightness of neighboring pixels \cite{pfister2011recognising}. In the realm of deep learning, convolutional and pooling layers are employed to extract spatial features, while recent advancements incorporate attention mechanisms to enhance the network’s ability to focus on critical regions by generating weighted feature map masks. Additionally, Graph Convolutional Networks (GCNs) model facial AUs as graph nodes, further optimizing spatial feature extraction by leveraging prior knowledge to improve performance \cite{xie2020assisted}. Given the transient and spontaneous nature of MEs, temporal feature extraction is equally critical. Traditional handcrafted feature methods often treat video data as three-dimensional spatio-temporal volumes, applying feature extraction operators, such as LBP-TOP \cite{sun2020multi} and 3DHOG \cite{polikovsky2009facial}, across sections encompassing both spatial and temporal dimensions. Optical flow-based features achieve similar objectives by integrating local spatial and temporal information \cite{liong2019shallow, xia2019spatiotemporal}. In deep learning approaches, preprocessed optical flow matrices are commonly used as input instead of raw images to capture temporally proximate information \cite{liu2019neural, kumar2021micro}. To model longer-term temporal dependencies, some studies treat video sequences as three-dimensional matrices or employ Recurrent Neural Networks (RNNs) and Long Short-Term Memory (LSTM) networks for temporal modeling \cite{khor2018enriched, kim2016micro}. These approaches significantly enhance the robustness and accuracy of ME feature extraction.

\textbf{MER based on multi-modal fusion.} Benlamine \textit{et al.} \cite{benlamine2016physiology} captured MEs and corresponding EEG signals from participants viewing image stimuli, employing time- and frequency-domain features of 1 s EEG signals to train machine learning algorithms for MER. Kim \textit{et al.} \cite{kim2022classification} classified discrete emotions in MEs using EEG and facial electromyography (EMG) signals, offering electrode placement guidelines for designing wearable MER devices. Saffaryazdi \textit{et al.} \cite{saffaryazdi2022emotion} integrated EEG, GSR, and PPG to detect emotions in MEs, demonstrating that combining MEs with brain and peripheral physiological signals enhances the reliability of detecting underlying emotions, underscoring the distinct advantage of MEs over MaEs in revealing authentic emotions. Zhao et al. \cite{zhao2022responses} utilized EEG to investigate the reorganization of functional brain networks during MEs, aiming to uncover neural mechanisms that could provide electrophysiological indicators for MER. Their findings revealed that during MEs, participants exhibited higher global and nodal efficiency in the frontal, occipital, and temporal regions, confirming the possibility of using EEG to recognize MEs. Furthermore, they conducted an EEG study on the differences in brain activation between MEs and MaEs, exploring the neural mechanisms underlying MEs and their distinctions from MaEs from a neuroscience perspective \cite{zhao2024micro}. The results showed that under positive emotions, MEs significantly activated the premotor cortex, supplementary motor cortex, and middle frontal gyrus, providing a theoretical foundation for multimodal MER.

\subsubsection{ME Spotting Methods}

The objective of ME spotting is to precisely identify the onset, apex and offset frames of MEs in video sequences. However, the transient and subtle nature of MEs presents substantial challenges for precise keyframe localization, resulting in poor performance of vision-based spotting algorithms and difficulty meeting practical requirements.

\textbf{ME Spotting based on visual information.} ME spotting methods based on visual information primarily involves two approaches: handcrafted feature-based and deep learning-based. Handcrafted-based methods typically rely on detailed feature engineering and signal processing design. These approaches extract features such as LBP, HOG, HOOF and MDMO. They identify the frame with the highest feature magnitude as the apex frame or determine the expression interval using threshold strategies. For instance, Han \textit{et al.} \cite{han2018cfd} combined LBP and MDMO to extract more robust features, where LBP captures texture information, and MDMO detects motion features by identifying the dominant direction in optical flow histograms. Esmaeili \textit{et al.} \cite{esmaeili2020automatic} developed an enhanced LBP descriptor that computes LBP across fifteen planes to capture critical information related to MEs. Although this approach achieved promising results, it is computationally intensive and time-consuming. More recently, Ma \textit{et al.} \cite{ma2017region} introduced a feature extraction method based on the Regional Histogram of Oriented Optical Flow (RHOOF) to automatically detect apex frames, achieving robust results. However, since handcrafted feature methods heavily depend on manually designed feature extractors, they require extensive expertise and involve complex parameter tuning processes. With the rapid development of deep learning technologies, several well-designed network architectures, such as AlexNet \cite{pan2020local}, ResNet \cite{tran2021micro}, and Long Short-Term Memory (LSTM) networks \cite{wang2021mesnet}, have gradually been applied to ME Spotting. These approaches, which leverage the advantages of automated feature engineering and end-to-end training, are becoming mainstream in the field. Zhang \textit{et al.} \cite{zhang2018smeconvnet} were the first to introduce deep learning into ME Spotting. They used Convolutional Neural Networks (CNNs) to extract deep features and proposed a feature matrix processing method to search for apex frames from long video sequences. Verburg \textit{et al.} \cite{verburg2019micro} were the first to apply RNNs to ME Spotting. Their method extracts HOOF features from sliding windows and inputs these features into an RNN composed of LSTM units for classification. However, the brief duration and subtle intensity of MEs continue to pose challenges for improving the accuracy and robustness of deep learning frameworks.

\textbf{ME spotting based on multi-modal fusion.} When individuals experience intense emotions, sympathetic nervous system activity is expected to increase. During such moments, MEs may briefly appear, accompanied by changes in ECG signals, such as fluctuations in heart rate (HR) \cite{giannakakis2019review}. MEs and HR variations represent external behavioral expressions and internal physiological responses to emotions, respectively, both influenced by emotional triggers and regulatory mechanisms. Early research estimated HR from facial videos by analyzing subtle color changes and micro-movements caused by cardiovascular pulsations. For example, Gupta \textit{et al.} \cite{gupta2018exploring} investigated the feasibility of using estimated instantaneous HR fluctuations to identify MEs. Their findings revealed that HR fluctuation amplitudes significantly increased during the occurrence of MEs. Building on this insight, they used variations in instantaneous HR to distinguish genuine ME intervals from the plausible ME intervals. Similarly, Zhang \textit{et al.} \cite{zhang2022your} detected participants’ HR from ME videos and developed a spatiotemporal fusion network incorporating HR information. Their results demonstrated that integrating image data with HR signals via a multimodal learning approach provides a more comprehensive representation of ME features. Additionally, Setyaningrum \textit{et al.} \cite{setyaningrum2024integrating} conducted real-time detection of MEs and HR from facial videos to explore their relationship. Using Eulerian Video Magnification (EVM) \cite{shahadi2020eulerian} and Fast Fourier Transform (FFT) \cite{rajaby2022structured} for HR detection, they found that emotions reflected by MEs influenced HR: angry expressions corresponded to higher HRs, sad expressions to lower HRs, and happy expressions exhibited greater variability. Despite these advancements, most studies estimated HR fluctuations from facial videos, which are susceptible to environmental disturbances. To address this limitation, Zou \textit{et al.} \cite{zou2022concordance} collected facial MEs and ECG signals, and their statistical analysis revealed a significant correlation between MEs and the time-domain features of heart rate variability (HRV). However, their study did not extend to leveraging HRV for the recognition or spotting of MEs. 

    \begin{figure*}[b]
        \centering
        \includegraphics[width=0.95\linewidth]{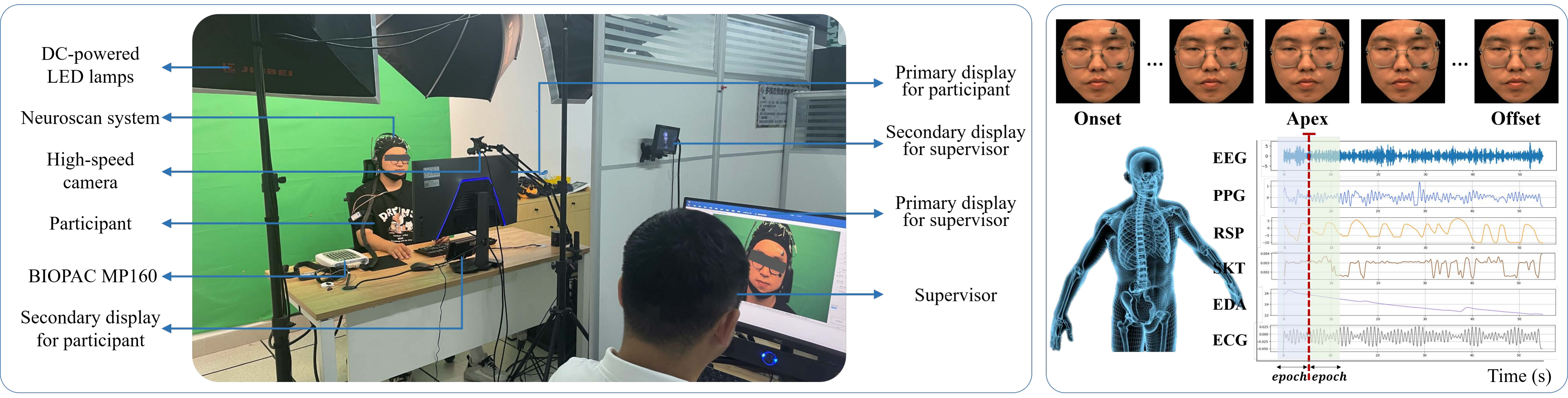}
        \caption{Experimental setup and signal examples of the MMME dataset acquisition system. The left panel depicts the laboratory environment, including the participant and data acquisition equipment. The right panel presents representative samples of the collected multimodal data, featuring key frames of ME and time-series representations of multiple PS.}
        \label{fig:fig-Experiment}
    \end{figure*}

\section{MMME Database Profile}

In this study, we developed a comprehensive multimodal ME dataset, named MMME, which synchronously captures facial action signals (MEs), central nervous system signals (EEG), and multiple peripheral physiological signals (PERI-PS) including PPG, RSP, SKT, EDA, and ECG. These modalities serve as reliable indicators of emotional states, providing robust data support for exploring the integration of multi-source data to enhance the performance of emotion recognition tasks. Each sample was meticulously annotated with facial AUs and emotion category labels. Furthermore, considering the co-occurrence of MEs and macro-expressions (MaEs) during video viewing by participants, the MMME dataset also systematically collected and annotated MaEs, thereby enriching the dataset's diversity and offering a more comprehensive foundation for research on emotional representation. The following sections detail the dataset construction process, including participants and experiments, emotion elicitation materials, experimental paradigm, annotation methods, and statistical analysis of the collected samples.

\subsection{Participants and Equipments}
\label{Equipment}

In this study, we recruited a total of 75 participants, including 44 males and 31 females, with ages ranging from 21 to 40 years. The average age was 28.80 years (standard deviation = 4.55). All participants were right-handed, had normal or corrected vision, had no history of mental illness, and were not at risk for depression. Prior to the experiment, all participants provided written informed consent and consented to the use of their likeness. Each participant exhibited at least one ME in the video clips.

The experimental setup, as illustrated in the left panel of Fig. \ref{fig:fig-Experiment}, was conducted in a controlled environment. To eliminate 50 Hz AC-induced flickering, we implemented a specialized lighting system consisting of three 600-watt DC-powered LED lights equipped with umbrella reflectors. These lights were strategically positioned to focus illumination on the participant's face, ensuring consistent and high-intensity lighting conditions. Specifically, the lighting configuration comprised one central light positioned directly in front of the participant and two additional lights placed at 45 angles to the left and right of the participant. Participants were seated one meter from the main display monitor, with a green background panel positioned behind them to facilitate subsequent facial image processing.
The participants' facial images were acquired using a high-speed camera system (XIMEA MQ022CG-CM, Germany) equipped with an industrial lens (KOWA LM16JC10M, 16 mm focal length, Japan) operating at 150 fps with a resolution of 2,040 $\times$ 1,080 pixels. The camera was positioned above the main display, maintaining direct alignment with participants' facial orientation. Neural activity was recorded through a 64-channel QuickCap coupled with the Neuroscan SynAmps RT system, acquiring EEG data at 1000 Hz. Concurrently, PERI-PS were captured using the BIOPAC MP160 acquisition system and AcqKnowledge 5 software (BIOPAC Systems, Inc., USA) at an identical sampling rate of 1000 Hz. For ECG recording, Ag/AgCl disposable vinyl electrodes (EL503; BIOPAC Systems, Inc.) and conductive gel were used to collect signals in a lead II configuration. EDA data were collected from the inner phalanges of the index and middle fingers of the left hand. A shared trigger between the camera, Neuroscan, and BIOPAC system was employed to generate timestamps, ensuring precise synchronization between facial image capture and physiological signal acquisition.

\subsection{Emotion Elicitation Materials}

Given that video stimuli containing both visual and auditory elements can effectively elicit significant emotional states and physiological responses in participants within laboratory settings \cite{polo2023comparative}, we curated a dataset of 320 video clips from online sources to construct an emotion-inducing stimulus library. To ensure the immediacy and intensity of emotional responses, each clip was edited to a duration of no more than one minute. These videos were designed to evoke seven distinct emotions: happiness, surprise, sadness, fear, anger, disgust, and contempt. These emotions are considered to be universal, independent of cultural, historical, or individual differences, and are typically expressed in a similar manner across individuals. To assess the effectiveness of these videos in eliciting emotions, we recruited 20 independent evaluators to watch and rate the clips. After viewing randomly presented clips, the evaluators completed a questionnaire based on a nine-point Likert scale \cite{jebb2021review}, rating each video according to three criteria: arousal, valence, and emotion category. Notably, these evaluators did not participate in the subsequent emotion induction experiment to prevent potential bias.
Conducting a reliability assessment of the questionnaire is a crucial step in ensuring data quality and enhancing the reliability and validity of research conclusions. We utilized Cronbach's alpha coefficient \cite{tavakol2011making} to assess the internal consistency of the questionnaire results, defined as follows:

    \begin{equation}
        \alpha=\frac{n}{n-1}\left(1-\frac{\sum s_{i}^{2}}{s_{T}^{2}}\right),
    \end{equation}
where $n$ is the number of items, $s_{i}^{2}$ is the variance of the $i^{th}$ item, and $s_{T}^{2}$ is the variance of the total scores. The calculated Cronbach's alpha coefficients for the questionnaire exceeded 0.80, indicating good internal consistency and trustworthiness of the results. 

Furthermore, we employed the Intraclass Correlation Coefficient (ICC) \cite{mehta2018performance} to assess consistency, which is a widely used reliability metric in the analysis of assessor reliability. The formula is as follows:

    \begin{equation}
        \rho=\frac{MS_{r}-MS_{e}}{MS_{r}+(n-1) MS_{e}},
    \end{equation}
where $MS_{r}$ is the mean square between subjects, $MS_{e}$ is the mean square error, and $n$ is the number of subjects. At a 95\% confidence interval, the ICC coefficients for pleasure and arousal were 0.768 and 0.697, respectively, indicating good consistency for both dimensions.

Based on the evaluation results, we ultimately selected six representative video clips for each emotional category, resulting in a total of 42 video stimuli being utilized in the experiment. These carefully curated video materials contain intensive stimulation points with high arousal potential, effectively eliciting singular target emotions.

\subsection{Experimental Paradigm}
\label{Procedure}

\begin{figure}[b]
        \centering
        \includegraphics[width=1\linewidth]{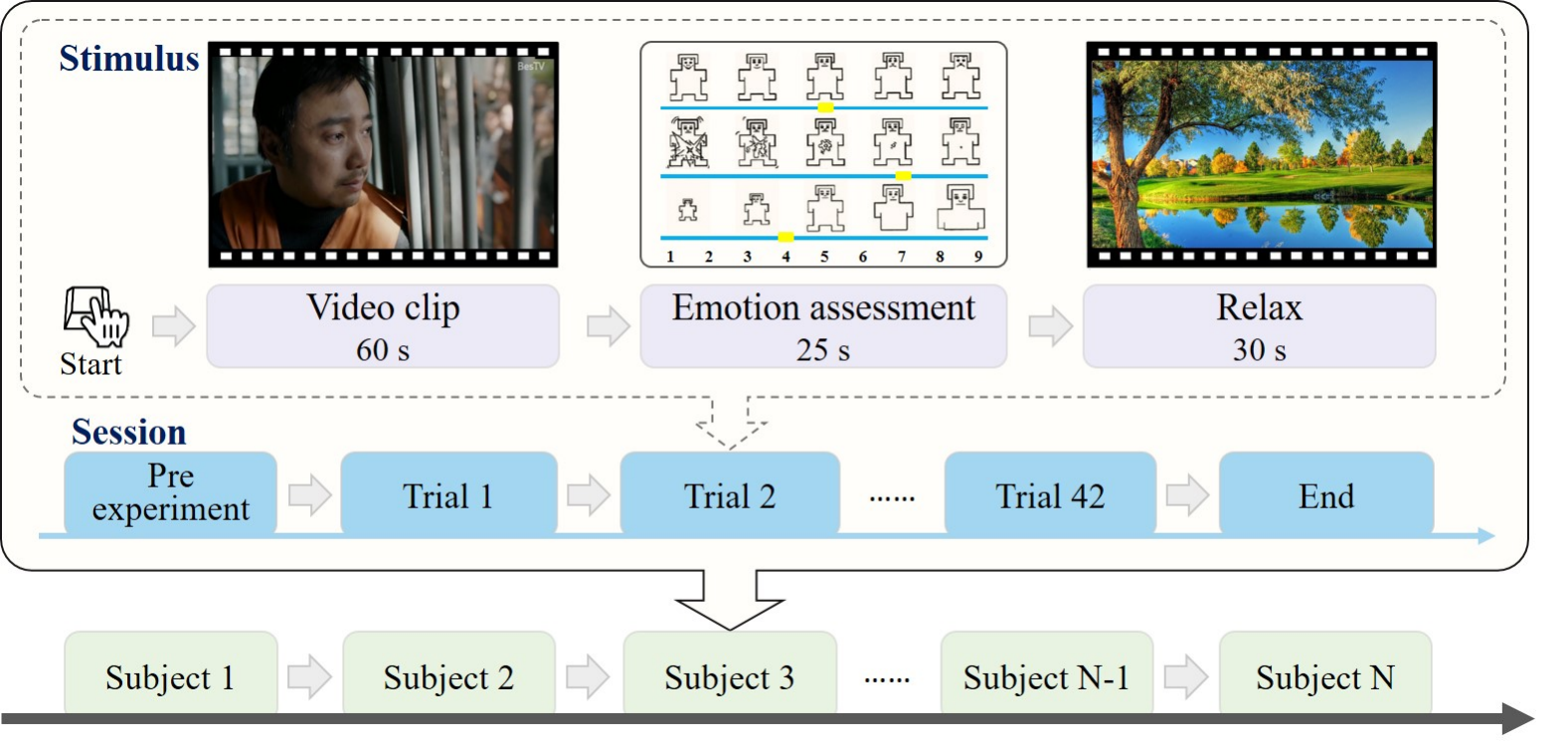}
        \caption{The detail of the experimental procedure. Our experiment included a total of 42 trials. Throughout the experiment, facial videos and various physiological signals were recorded simultaneously.}
        \label{fig:Procedure}
    \end{figure}

To familiarize participants with the experimental procedure, a pre-experiment was conducted prior to the formal experiment. The videos used in the pre-experiment were not repeated in the formal experiment, and the data collected during the pre-experiment were not included in the final analysis. In the formal experiment, participants controlled the playback of the video clips by pressing the space bar, with each clip lasting approximately 60 s. After each video segment, participants were required to quickly and accurately report the type of emotion, emotional valence, and arousal level within 25 s. A 30 s neutral video was then played to help participants relax. This sequence constituted a complete trial. Our experiment included a total of 42 trials, meaning each participant watched 42 video clips. Throughout the experiment, facial videos and various physiological signals were recorded simultaneously. Additionally, participants were instructed to minimize head and body movements as much as possible. Fig. \ref{fig:Procedure} provides a detailed overview of the experimental process.

    \begin{table*}[t]
        \centering
        \caption{The list of AU codes involved in MEs in the MMME dataset, along with their corresponding action descriptions, occurrence regions, and counts}
        \label{tab:AUs}
        \large
        \renewcommand{\arraystretch}{1.2}
        \resizebox{0.95\textwidth}{!}{%
        \begin{tabular}{cccccccccccc}
        \hline
        \toprule[0.5pt]
        \textbf{AUs} & \textbf{Action description} & \textbf{Region} & \textbf{Count} & \textbf{AUs} & \textbf{Action description} & \textbf{Region} & \textbf{Count} & \textbf{AUs} & \textbf{Action description} & \textbf{Region} & \textbf{Count} \\
        \cmidrule(r){1-4} \cmidrule(r){5-8} \cmidrule(r){9-12}
        AU1 & Inner Brow Raiser & Upper & 54 & AU10 & Upper Lip Raiser & Mid & 33 & AU17 & Chin Raiser & Lower & 39 \\
        AU2 & Outer Brow Raiser & Upper & 24 & AU11 & Nasolabial Deepener & Mid & 6 & AU18 & Lip Puckerer & Lower & 8 \\
        AU4 & Brow Lowerer & Upper & 357 & AU12 & Lip Corner Puller & Mid & 48 & AU20 & Lip streteher & Lower & 15 \\
        AU5 & Upper Lid Raiser & Mid & 12 & AU13 & Cheek Puffer & Lower & 12 & AU25 & Lips parted & Lower & 12 \\
        AU6 & Cheek Raiser & Upper & 17 & AU14 & Dimpler & Lower & 165 & AU26 & Jaw Drop & Lower & 10 \\
        AU7 & Lid Tightener & Mid & 54 & AU15 & Lip Corner Depressor & Lower & 23 & AU41 & Lid droop & Mid & 11 \\
        AU9 & Nose Wrinkler & Mid & 12 & AU16 & Lower Lip Depressor & Lower & 16 & AU42 & Slit & Upper & 8 \\ \hline
        \toprule[0.5pt]
        \end{tabular}%
        }
            \begin{tablenotes} 
                \fontsize{8pt}{8pt}\selectfont 
                \item[1] $^1$ Upper: the upper face; Mid: the mid-face; Lower: the lower face.
            \end{tablenotes}
    \end{table*}

Unlike traditional ``immersive emotion induction'' parad-igms, this study aims to establish an experimental environment specifically tailored for ME research. Given the intrinsic nature of MEs as ``suppressed-leakage'' phenomena \cite{ekman2003darwin}, it is essential to maintain a moderate level of facial control pressure throughout the experiment. Although the classic ME elicitation paradigm instructs participants prior to the experiment to suppress facial expressions and minimize bodily movements \cite{yan2014casme,qu2017cas}, our observations revealed that participants still exhibited ``uncontrolled'' behaviors, such as unconscious relaxation of facial control and frequent head or body movements. Notably, these phenomena persisted despite participants being informed that such behaviors would result in a reduction of their compensation, and they often required considerable time to return to the required suppression state. This finding suggests that pre-experimental instructions and intrinsic motivation alone are insufficient to ensure consistent experimental conditions.

To address this issue, we developed the Continuous Monitoring and Real-time Reminding (CMRR) paradigm, in which the experimenter continuously monitors participants' facial expressions and body movements in real time. When a participant exhibits signs of facial relaxation or frequent head and body movements, the experimenter triggers a visual prompt on a small auxiliary display. This prompt serves as an immediate signal for the participant to swiftly restore the required level of facial suppression. Experimental results demonstrate that the CMRR paradigm not only accelerates participants’ recovery to the required state but also significantly enhances their ability to maintain this state in subsequent trials. From a theoretical perspective, the CMRR paradigm aligns closely with the core feature of MEs as ``emotional leakage within a controlled environment'' \cite{ben2021video}. In contrast to fully immersive paradigms, CMRR achieves a balance between experimental control and ecological validity by minimizing interference—cues are triggered only when necessary.

\subsection{ME Annotation}

Following the completion of 2,814 trials and the collection of nearly 80 hours of multimodal emotional data, we rigorously adhered to the annotation protocols established by authoritative ME datasets, such as the CASME series \cite{yan2013casme, yan2014casme, qu2017cas, li2022cas}. The annotation of ME samples was conducted independently by two certified experts in the FACS. Using a custom-developed annotation software, the experts analyzed each video frame by frame to precisely identify the onset, apex, and offset frames of each expression sequence. In this study, facial expressions meeting either of the following criteria were classified as ME samples: (1) total duration from onset to offset not exceeding 500 ms, or (2) duration from onset to apex frame not exceeding 250 ms. Each identified ME sample was subsequently annotated with corresponding AUs and categorized into one of seven basic emotion categories: \textit{happiness}, \textit{surprise}, \textit{sadness}, \textit{fear}, \textit{anger}, \textit{disgust}, or \textit{contempt}. Furthermore, to assess inter-coder reliability, we calculated consistency score ($r$) between the two annotators, with the computational methodology detailed as follows:

    \begin{equation}
        r=2 \times \frac{AU\left(A_{1}\right) \cap AU\left(A_{2}\right)}{All_{AU}},
    \end{equation}
where $AU\left(A_{1}\right) \cap AU\left(A_{2}\right)$ is the number of AUs both annotators agreed, and $All_{AU}$ is the total number of AUs in an ME labeled out by the two annotators. In this study, the inter-annotator reliability score was 0.84, which is comparable to those reported for the CAS(ME)$^2$ (0.82) \cite{qu2017cas} and DFME (0.83) \cite{zhao2023dfme} datasets. Ultimately, through comprehensive discussions, the annotators reached a consensus, producing a consistent and reliable set of annotations.

    \begin{figure*}[b]
        \centering
        \includegraphics[width=0.95\linewidth]{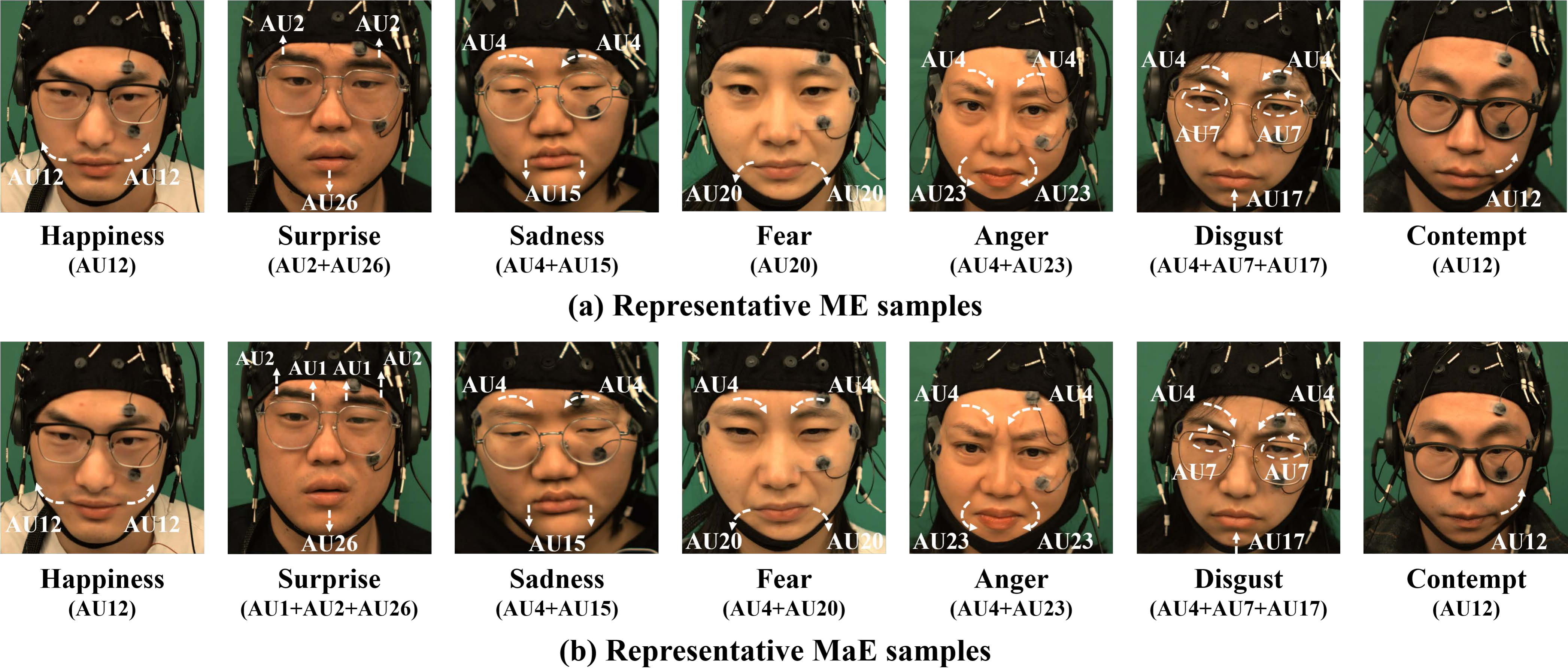}
        \caption{Representative ME and MaE samples for seven emotion categories in MMME. White arrows indicate the locations and movement patterns of key AUs, and the circular arrows in the eyelid region indicate the orbital narrowing motion caused by the contraction of the orbicularis oculi muscle.}
        \label{fig:sample}
    \end{figure*}

Drawing upon commonly observed AU patterns in published ME datasets and considering participants' actual performance, we annotated a total of 21 distinct AU categories. These AUs were distributed across facial regions as follows: 5 categories in the upper face (from hairline to below eyebrows), 7 categories in the mid-face (from below eyebrows to nasal base), and 9 categories in the lower face (from nasal base to chin). Table \ref{tab:AUs} presents a detailed list of AU codes, along with their corresponding facial action descriptions, regions, and counts. It can be observed that the three most frequently occurring AUs are AU1 (Inner Brow Raiser), AU2 (Outer Brow Raiser), and AU12 (Lip Corner Puller). These AUs are primarily located in facial regions most relevant to MEs: the eyebrows and the corners of the mouth, which are critical for ME analysis.

After meticulous annotation, we obtained 634 MEs (mean duration = 389 ms, SD = 97 ms), 2841 MaEs (mean duration = 1634 ms, SD = 328 ms) and multimodal physiological data for 4,200 trials. Fig. \ref{fig:sample} illustrates representative samples of MEs and MaEs across seven emotion categories in our MMME dataset, accompanied by their corresponding AUs. As shown, MEs are notably more subtle than MaEs, rendering their detection and recognition considerably more challenging. It is worth noting that, the MaEs collected in our dataset demonstrate greater restraint, characterized by reduced motion amplitudes and shorter durations, when compared to those in datasets obtained through emotion-elicitation paradigms (\textit{e.g.}, FER+ \cite{barsoum2016training}, BP4D+ \cite{zhang2014bp4d}, CK+ \cite{lucey2010extended}). This distinction arises from the neutral paradigm employed in our study, where participants were explicitly instructed to suppress facial movements during data collection. This unique feature of the dataset offers significant value for investigating facial expressions under suppression, particularly in scenarios or cultural contexts where subtle or restrained emotional expression is prevalent.

\subsection{Data Statistics and Analysis}

    \begin{figure}[!t]
        \centering
        \includegraphics[width=1\linewidth]{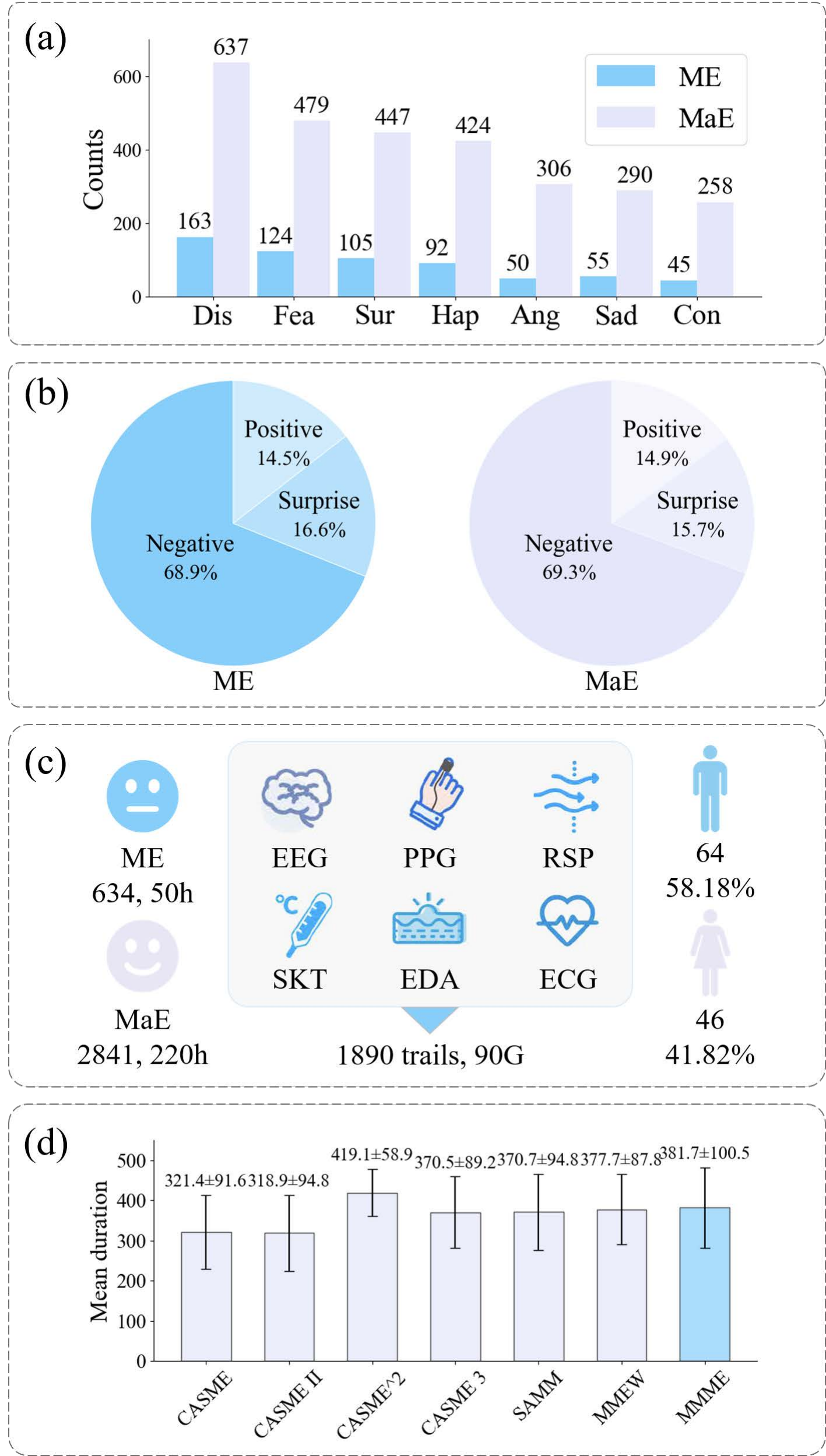}
        \caption{Statistical analysis of annotated samples in the MMME dataset.}
        \label{fig:data}
    \end{figure}

In this section, we perform a statistical analysis of the annotated samples. The results, presented in Fig. \ref{fig:data}, highlight four key characteristics: (a) the distribution of seven basic discrete emotions; (b) the distribution of three composite emotions (``Negative,'' ``Positive,'' and ``Surprise''); (c) the dataset size and participant gender composition; and (d) a comparison of mean ME durations and their standard deviations across benchmark datasets. The statistical analysis reveals that the number of ME samples is approximately one-fourth that of MaEs (634 vs. 2,841). This observation underscores the inherent difficulty of eliciting MEs in experimental settings due to their spontaneous and transient nature. Regarding the distribution of emotion categories, a pronounced imbalance is evident---an issue commonly observed in publicly available ME datasets. Specifically, samples corresponding to ``Disgust,'' ``Fear,'' and ``Surprise'' are relatively abundant, whereas those for ``Sadness'' and ``Contempt'' are significantly underrepresented. This disparity may stem from two primary factors: (1) variations in the difficulty of eliciting different emotions in experimental paradigms and (2) differences in the recognition difficulty of specific facial AUs. Notably, among the seven basic emotions, five (``Anger,'' ``Disgust,'' ``Fear,'' ``Sadness,'' and ``Contempt'') fall under the ``Negative'' valence category, leading to a predominance of ``Negative'' samples within the composite emotion distribution. In terms of data modality, the novelty of the MMME dataset lies in its multimodal data acquisition framework, which simultaneously records facial expression videos, central nervous system signals, and peripheral nervous system signals. This comprehensive approach provides a robust foundation for modeling the relationship between facial expressions and neural activity. Duration analysis indicates that the average ME duration in the MMME dataset (381.7 ± 100.5 ms) is consistent with those observed in publicly available datasets such as the CASME series, SAMM, and MMEW. This consistency validates the reliability of our dataset in terms of temporal distribution characteristics. In summary, through its multimodal nature and standardized annotation framework, the MMME dataset not only addresses the limitations of existing unimodal ME datasets but also serves as a crucial experimental platform for investigating the neural mechanisms underlying MEs and advancing their decoding algorithms.

    \begin{figure}[b]
        \centering
        \includegraphics[width=0.95\linewidth]{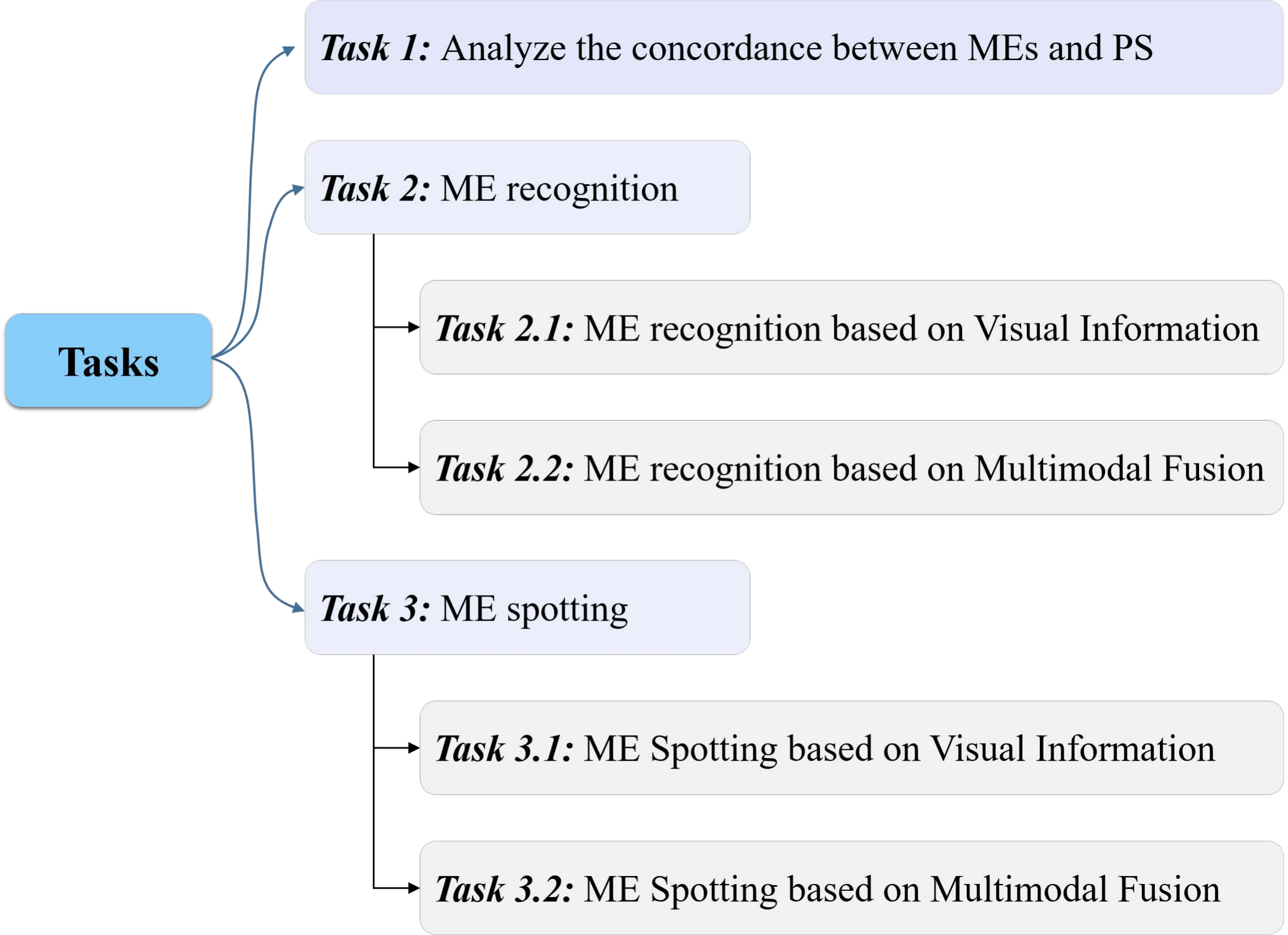}
        \caption{Task framework for dataset evaluation: concordance, recognition, and spotting of MEs.}
        \label{fig:Tasks}
    \end{figure}
    
\section{Database Evaluation}

In this section, we design a series of experimental tasks to evaluate the effectiveness of the MMME dataset, with the task framework illustrated in Fig. \ref{fig:Tasks}. The evaluation tasks primarily focus on three aspects: (1) Concordance analysis between MEs and PS, (2) ME recognition (categorized into vision-based methods and multimodal fusion approaches), and (3) ME spotting (similarly divided into two approaches). The concordance validation between MEs and PS provides a theoretical foundation for designing fusion strategies, thereby effectively leveraging the complementary advantages across different modalities, as detailed in Section \ref{subsection-concordance}. Both ME recognition (Section \ref{subsection-recognition}) and ME spotting (Section \ref{subsection-spotting}) tasks incorporate unimodal and multimodal analyses, establishing comprehensive benchmarks for dataset validation.

\subsection{The Concordance Between MEs and PS}
\label{subsection-concordance}

Numerous studies have demonstrated that multimodal fusion techniques offer significant advantages in the field of affective computing, substantially improving the accuracy and robustness of emotion recognition \cite{kim2024enhancing, tang2024hierarchical, han2024fmfn}. However, research in the specific domain of ME analysis remains predominantly confined to unimodal visual approaches. Notwithstanding this limitation, recent years have witnessed emerging attempts to integrate MEs with physiological signals to more comprehensively capture subtle emotional variations. For instance, Kim \textit{et al.} \cite{kim2022classification} successfully classified discrete emotions in facial MEs by fusing EEG and facial EMG data. Saffaryazdi \textit{et al.} \cite{saffaryazdi2022emotion} employed a multimodal approach incorporating EEG, GSR, and PPG to identify emotions underlying MEs. Furthermore, Zhao \textit{et al.} \cite{zhao2024micro} conducted a focused investigation comparing neural activation patterns between MEs and MaEs using EEG data, revealing differences in brain activity. Despite these pioneering efforts that have expanded the possibilities of multimodal research, they share a common limitation: none have systematically examined the concordance or underlying mechanisms between MEs and multimodal physiological data. This knowledge gap results in a lack of theoretical foundation for multimodal fusion strategies, leaving the complementary advantages among modalities insufficiently explored and utilized.
From a neurophysiological perspective, MEs and PS play complementary roles in emotional expression. During intense emotional episodes, MEs reflect transient emotional changes through subtle facial muscle movements, while PS record dynamic activation patterns of the autonomic nervous system. Theoretically, effective integration of these modalities could not only provide a more comprehensive emotional profile by capturing both external behavioral manifestations (MEs) and internal physiological responses, but might also reveal synergistic patterns between them. Such integration would thereby facilitate the development of more precise emotion recognition models.

To optimize the fusion of visual signals and physiological features during ME occurrences, this study explores the concordance and potential interaction mechanisms between MEs and various PERI-PS under high-arousal emotional stimuli. Specifically, we investigated whether these modalities exhibit synchronous variations during emotional responses. Since PERI-PS were continuously recorded while MEs occur sparsely, the apex frames of MEs were utilized to segment the corresponding PS. Considering that physiological responses to emotional stimuli typically exhibit temporal delays, an appropriate redundancy was introduced in the signal segmentation process to cover this delay. Research has indicated that emotional fluctuations can be reflected in PERI-PS within a temporal window ranging from 3 to 15 s. For instance, Kim \textit{et al.} \cite{kim2007bimodal} performed an emotion induction experiment, simultaneously recording multiple signals, with annotated segments ranging from 3 to 15 s in duration. Other studies \cite{gunes2010automatic, jerritta2011physiological} have also confirmed durations associated with emotional changes. Therefore, the duration of 3 s and 15 s were chosen to cover the range of 3 to 15 s, and two additional durations (5 s, and 10 s) were also considered in our study. Consequently, this study employed epoch durations of 3, 5, 10, and 15 s centered around the apex frames of MEs for segmentation, while ensuring that no other MEs occurred within the same epoch. The open-source neurophysiological signal processing tool NeuroKit2\footnote{\url{https://neuropsychology.github.io/NeuroKit/}} was utilized for PERI-PS processing and feature extraction.

    \begin{figure}[t]
        \centering
        \includegraphics[width=1\linewidth]{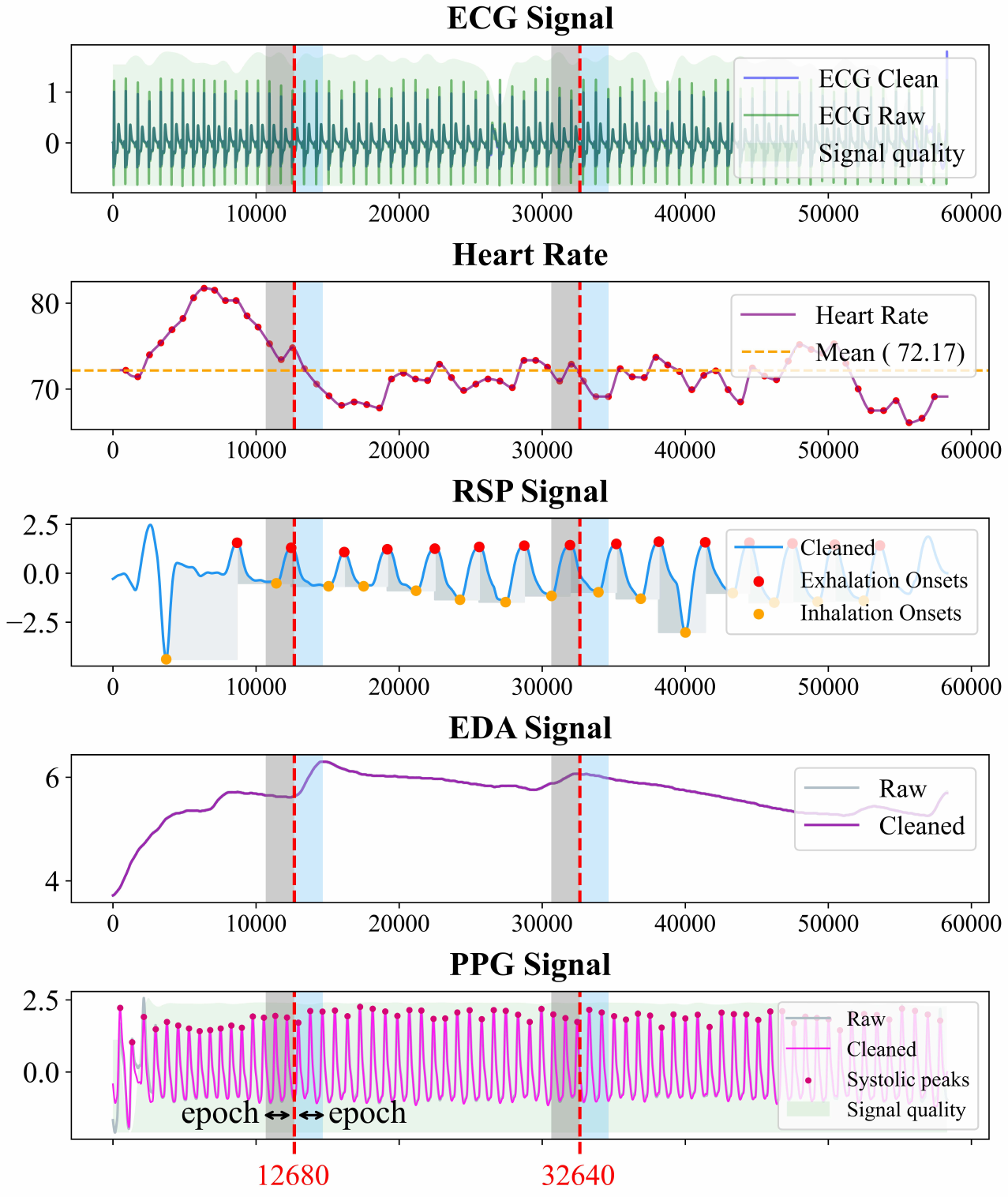}
        \caption{Temporal variations of different PERI-PS in a single trial, where two MEs occur with apex frames at 12,680 ms and 32,640 ms.}
        \label{fig:Signals}
    \end{figure}

Fig. \ref{fig:Signals} illustrates the temporal variations of different PS in a single trial. Vertical red dashed lines indicate the apex frames of two MEs in this trial, occurring at 12,680 ms and 32,640 ms. Gray-shaded regions denote the epochs before the apex frames, and blue-shaded regions represent the epochs after them. By examining significant differences in PERI-PS characteristics between these epochs, we aim to identify potential synchronization between MEs and PS. However, through visual inspection, there were no significant abrupt changes in the amplitude or phase of the signal before and after the occurrence of MEs. Therefore, we further extracted physiological features to investigate, in a statistically significant manner, whether there is consistency between MEs and PERI-PS features. Table \ref{tab: Paired t-test} presents the results of paired-sample t-tests, which were conducted to analyze the changes in various PERI-PS features before and after the onset of MEs across different time durations. The statistical measures reported include the mean, standard deviation, t-value, and p-value, with gray-shaded p-values indicating statistical significance at the $p < 0.05$ level.

The results reveal a significant association between the occurrence of MEs and certain physiological features. Specifically, within 3 s epochs, several HRV features showed significant differences, including HRV\_MeanNN ($p < 0.01$), HRV\_SDSD ($p < 0.01$), HRV\_MedianNN ($p < 0.01$), HRV\_Prc20NN ($p < 0.01$), HRV\_Prc80NN ($p < 0.01$), HRV\_MinNN ($p < 0.01$), HRV\_MaxNN ($p < 0.05$), and HRV\_TINN ($p < 0.05$). In addition, RSP features such as RSP\_Std\_dev ($p < 0.05$) and RSP\_Min\_value ($p < 0.05$), as well as EDA features including EDA\_Mean ($p < 0.05$) and EDA\_Mean\_amplitude ($p < 0.05$), also exhibited statistically significant changes. Similarly, within 5 s epochs, significant differences were observed in ECG features (RR\_Mean ($p < 0.05$), RR\_Std ($p < 0.05$)), HRV features (Heart\_Rate\_Mean ($p < 0.05$), HRV\_MeanNN ($p < 0.05$), HRV\_MedianNN ($p < 0.05$), HRV\_Prc20NN ($p < 0.05$), HRV\_Prc80NN ($p < 0.05$), HRV\_MinNN ($p < 0.05$)), RSP features (RSP\_Std\_dev ($p < 0.01$), RSP\_Min\_value ($p < 0.01$)), EDA features (EDA\_Num\_peaks ($p < 0.01$)), and SKT features (SKT\_Slope ($p < 0.01$)). Furthermore, several physiological features also showed significant differences before and after the occurrence of MEs in 10 s and 15 s epochs. For 10 s epochs, significant changes were identified in ECG features (RR\_Std ($p < 0.05$)), HRV features (Heart\_CVNN ($p < 0.05$), Heart\_CVSD ($p < 0.05$), Heart\_MedianNN ($p < 0.01$), HRV\_Prc20NN ($p < 0.05$)), RSP features (RSP\_Std\_dev ($p < 0.01$), RSP\_Median ($p < 0.05$), RSP\_Max\_value ($p < 0.01$), RSP\_Min\_value ($p < 0.05$)), EDA features (EDA\_Mean ($p < 0.01$), EDA\_Max ($p < 0.01$), EDA\_Min ($p < 0.01$)), and SKT features (SKT\_Max ($p < 0.05$)). Similar patterns were also found in the 15 s epochs. These findings provide evidence of a consistent and significant relationship between MEs and PS, thereby establishing a robust foundation for the proposed multimodal fusion of ME and physiological data in this study.

\subsection{ME Recognition Validation}
\label{subsection-recognition}

To evaluate the effectiveness of the MMME dataset, we conduct micro-expression recognition (MER) tasks in this study, classifying ME samples into emotion categories using both unimodal (visual-only) and multimodal (visual and physiological signal fusion) approaches. The initial recognition results establish a baseline performance, providing a reference for researchers to compare and assess the efficacy of different methods in future work.

\begin{landscape} 
    \begin{table}[h]
    \centering
    \large
    \caption{Paired sample t-test before and after the ME apex} 
    \label{tab: Paired t-test}
    \begin{threeparttable} 
    \renewcommand{\arraystretch}{1.15} 
    \resizebox{1.35\textheight}{!}{
    \begin{tabular}{cccccccccccccccccc}
    \hline
     & \textbf{Epoch} & \multicolumn{4}{c}{\textbf{3s}} & \multicolumn{4}{c}{\textbf{5s}} & \multicolumn{4}{c}{\textbf{10s}} & \multicolumn{4}{c}{\textbf{15s}} \\ \cmidrule(r){2-6} \cmidrule(r){7-10} \cmidrule(r){11-14} \cmidrule(r){15-18}
    \multirow{-2}{*}{\textbf{\begin{tabular}[c]{@{}c@{}}Physiological\\ Signals\end{tabular}}} & \textbf{Feature Name} & \textit{\textbf{pre.}} & \textit{\textbf{aft.}} & \textit{\textbf{t}} & \textit{\textbf{p}} & \textit{\textbf{pre.}} & \textit{\textbf{aft.}} & \textit{\textbf{t}} & \textit{\textbf{p}} & \textit{\textbf{pre.}} & \textit{\textbf{aft.}} & \textit{\textbf{t}} & \textit{\textbf{p}} & \textit{\textbf{pre.}} & \textit{\textbf{aft.}} & \textit{\textbf{t}} & \textit{\textbf{p}} \\ \hline
     & ECG\_Mean & -0.008 (0.005) & -0.009 (0.006) & 1.850 & 0.065 & -0.008 (0.003) & -0.008 (0.004) & 0.852 & 0.395 & -0.008 (0.002) & -0.008 (0.002) & 0.757 & 0.450 & -0.008 (0.002) & -0.008 (0.001) & 0.643 & 0.521 \\
     & ECG\_Std & 0.145 (0.075) & 0.146 (0.076) & -1.356 & 0.176 & 0.148 (0.075) & 0.148 (0.076) & -1.149 & 0.252 & 0.149 (0.076) & 0.149 (0.077) & -0.812 & 0.418 & 0.152 (0.075) & 0.152 (0.076) & -0.726 & 0.469 \\
     & ECG\_Max & 0.635 (0.289) & 0.634 (0.288) & 0.612 & 0.541 & 0.647 (0.290) & 0.647 (0.288) & -0.336 & 0.737 & 0.654 (0.299) & 0.655 (0.299) & -0.400 & 0.690 & 0.670 (0.294) & 0.670 (0.292) & 0.040 & 0.968 \\
     & ECG\_Min & -0.509 (0.324) & -0.510 (0.325) & 0.256 & 0.798 & -0.528 (0.329) & -0.529 (0.329) & 0.852 & 0.395 & -0.539 (0.329) & -0.539 (0.329) & 0.092 & 0.927 & -0.425 (0.210) & -0.401 (0.188) & -2.152 & {\cellcolor[HTML]{D3D3D3}0.035} \\
     & ECG\_Range & 1.145 (0.574) & 1.144 (0.571) & 0.238 & 0.812 & 1.175 (0.581) & 1.177 (0.577) & -0.802 & 0.423 & 1.194 (0.590) & 1.195 (0.589) & -0.267 & 0.789 & 1.105 (0.456) & 1.081 (0.432) & 2.197 & {\cellcolor[HTML]{D3D3D3}0.032} \\
     & RR\_Mean & -- & -- & -- & -- & 0.000 (0.000) & 0.000 (0.000) & 2.195 & {\cellcolor[HTML]{D3D3D3}0.032} & 0.000 (0.000) & 0.000 (0.000) & 0.904 & 0.368 & 0.000 (0.000) & 0.000 (0.000) & 2.251 & {\cellcolor[HTML]{D3D3D3}0.028} \\
     & RR\_Std & -- & -- & -- & -- & 0.000 (0.000) & 0.000 (0.000) & -2.078 & {\cellcolor[HTML]{D3D3D3}0.039} & 0.000 (0.000) & 0.000 (0.000) & -0.852 & {\cellcolor[HTML]{D3D3D3}0.040} & 0.000 (0.000) & 0.000 (0.000) & 0.831 & 0.407 \\
    \multirow{-8}{*}{ECG} & Heart\_Rate\_Mean & -- & -- & -- & -- & 72.461 (6.974) & 73.165 (6.649) & -2.324 & {\cellcolor[HTML]{D3D3D3}0.021} & 72.829 (6.679) & 72.894 (6.260) & -0.255 & 0.799 & 73.010 (6.926) & 73.019 (6.134) & -0.033 & 0.974 \\
     & HRV\_MeanNN & 834.627 (85.430) & 822.013 (79.389) & 3.333 & {\cellcolor[HTML]{D3D3D3}0.001} & 834.898 (76.870) & 825.213 (71.876) & 2.547 & {\cellcolor[HTML]{D3D3D3}0.012} & 830.045 (70.871) & 828.698 (68.417) & 0.424 & 0.672 & 828.160 (72.466) & 826.619 (65.152) & 0.457 & 0.648 \\
     & HRV\_SDNN & -- & -- & -- & -- & 32.823 (26.318) & 34.431 (31.447) & -0.704 & 0.482 & 41.363 (25.866) & 41.264 (23.489) & 0.050 & 0.961 & 44.098 (24.373) & 35.868 (16.187) & 2.626 & {\cellcolor[HTML]{D3D3D3}0.011} \\
     & HRV\_RMSSD & 34.649 (31.218) & 38.304 (44.704) & -1.116 & 0.265 & 38.116 (28.609) & 40.339 (34.848) & -0.903 & 0.367 & 39.595 (26.227) & 41.067 (26.973) & -0.705 & 0.481 & 39.235 (30.245) & 31.962 (14.419) & 2.251 & {\cellcolor[HTML]{D3D3D3}0.028} \\
     & HRV\_SDSD & 24.260 (22.159) & 41.422 (60.396) & -4.440 & {\cellcolor[HTML]{D3D3D3}0.000} & 39.640 (30.342) & 42.901 (39.090) & -1.190 & 0.235 & 40.907 (27.646) & 42.594 (28.334) & -0.765 & 0.445 & 40.079 (31.186) & 32.796 (14.957) & 2.195 & {\cellcolor[HTML]{D3D3D3}0.032} \\
     & HRV\_CVNN & 0.030 (0.025) & 0.036 (0.054) & -1.785 & 0.075 & 0.039 (0.032) & 0.042 (0.044) & -1.053 & 0.293 & 0.049 (0.028) & 0.042 (0.018) & 2.169 & {\cellcolor[HTML]{D3D3D3}0.033} & 0.055 (0.029) & 0.045 (0.018) & 2.665 & {\cellcolor[HTML]{D3D3D3}0.010} \\
     & HRV\_CVSD & 0.041 (0.035) & 0.048 (0.064) & -1.634 & 0.103 & 0.045 (0.034) & 0.049 (0.048) & -1.233 & 0.219 & 0.044 (0.031) & 0.039 (0.017) & 2.014 & {\cellcolor[HTML]{D3D3D3}0.047} & 0.049 (0.038) & 0.039 (0.015) & 2.165 & {\cellcolor[HTML]{D3D3D3}0.034} \\
     & HRV\_MedianNN & 834.733 (85.680) & 821.886 (83.234) & 3.168 & {\cellcolor[HTML]{D3D3D3}0.002} & 836.817 (76.884) & 827.719 (72.604) & 2.416 & {\cellcolor[HTML]{D3D3D3}0.016} & 787.995 (92.978) & 798.168 (92.370) & -2.770 & {\cellcolor[HTML]{D3D3D3}0.007} & 827.403 (72.934) & 826.386 (67.692) & 0.304 & 0.762 \\
     & HRV\_MadNN & 24.070 (23.778) & 24.896 (28.482) & -0.423 & 0.673 & 26.356 (19.568) & 26.604 (20.249) & -0.167 & 0.867 & 40.318 (23.713) & 39.268 (22.013) & 0.569 & 0.570 & 43.010 (24.941) & 44.374 (23.464) & -0.631 & 0.529 \\
     & HRV\_MCVNN & 0.029 (0.027) & 0.030 (0.037) & -0.781 & 0.436 & 0.031 (0.022) & 0.032 (0.025) & -0.525 & 0.600 & 0.049 (0.028) & 0.047 (0.026) & 0.600 & 0.549 & 0.052 (0.028) & 0.053 (0.027) & -0.664 & 0.508 \\
     & HRV\_IQRNN & 18.524 (16.469) & 23.107 (36.879) & -1.966 & 0.050 & 35.378 (36.621) & 37.476 (46.561) & -0.615 & 0.539 & 53.160 (33.479) & 51.459 (29.749) & 0.687 & 0.493 & 57.373 (33.103) & 57.560 (29.705) & -0.069 & 0.945 \\
     & HRV\_SDRMSSD & 0.732 (0.090) & 0.742 (0.100) & -1.634 & 0.103 & 0.888 (0.242) & 0.878 (0.235) & 0.557 & 0.578 & 1.109 (0.382) & 1.067 (0.344) & 1.351 & 0.178 & 1.189 (0.394) & 1.171 (0.321) & 0.523 & 0.602 \\
     & HRV\_Prc20NN & 823.523 (85.214) & 808.261 (86.755) & 3.445 & {\cellcolor[HTML]{D3D3D3}0.001} & 813.172 (77.488) & 801.512 (82.673) & 2.366 & {\cellcolor[HTML]{D3D3D3}0.019} & 759.933 (90.295) & 769.739 (89.623) & -2.464 & {\cellcolor[HTML]{D3D3D3}0.016} & 793.340 (68.306) & 792.313 (62.532) & 0.290 & 0.772 \\
     & HRV\_Prc80NN & 845.751 (86.809) & 835.757 (77.662) & 2.794 & {\cellcolor[HTML]{D3D3D3}0.006} & 856.791 (81.853) & 847.794 (73.769) & 2.439 & {\cellcolor[HTML]{D3D3D3}0.015} & 862.885 (78.437) & 860.680 (74.233) & 0.624 & 0.533 & 863.771 (81.348) & 862.277 (72.718) & 0.392 & 0.696 \\
     & HRV\_pNN50 & 10.349 (19.741) & 11.131 (20.050) & -0.550 & 0.583 & 15.020 (19.105) & 13.526 (17.879) & 1.153 & 0.250 & 16.783 (16.827) & 17.017 (16.815) & -0.217 & 0.828 & 16.310 (15.612) & 18.295 (16.240) & -1.883 & 0.062 \\
     & HRV\_pNN20 & 30.144 (24.957) & 30.054 (25.251) & 0.046 & 0.963 & 42.410 (23.413) & 43.705 (22.488) & -0.733 & 0.464 & 49.753 (21.426) & 52.219 (19.354) & -1.789 & 0.075 & 50.095 (20.420) & 52.475 (18.722) & -1.641 & 0.103 \\
     & HRV\_MinNN & 816.058 (85.715) & 799.509 (91.344) & 3.416 & {\cellcolor[HTML]{D3D3D3}0.001} & 795.171 (78.586) & 783.908 (85.448) & 2.061 & {\cellcolor[HTML]{D3D3D3}0.040} & 768.469 (75.479) & 764.263 (78.021) & 0.806 & 0.421 & 752.027 (82.696) & 745.047 (79.110) & 0.929 & 0.354 \\
     & HRV\_MaxNN & 853.101 (88.191) & 844.560 (77.854) & 2.376 & {\cellcolor[HTML]{D3D3D3}0.018} & 871.713 (87.160) & 864.116 (74.724) & 1.943 & 0.053 & 896.330 (92.453) & 892.608 (81.440) & 0.837 & 0.404 & 905.007 (96.092) & 905.597 (84.897) & -0.112 & 0.911 \\
     & HRV\_HTI & 2.076 (0.505) & 2.117 (0.615) & -0.918 & 0.359 & 3.367 (1.083) & 3.311 (1.202) & 0.562 & 0.575 & 4.994 (1.757) & 4.883 (1.537) & 0.694 & 0.488 & 6.076 (2.033) & 6.031 (1.685) & 0.209 & 0.835 \\
    \multirow{-19}{*}{HRV} & HRV\_TINN & 0.000 (0.000) & 0.000 (0.000) & -1.501 & {\cellcolor[HTML]{D3D3D3}0.030} & 10.209 (31.146) & 10.894 (39.131) & -0.230 & 0.818 & 41.455 (49.042) & 51.398 (62.732) & -1.915 & 0.057 & 63.968 (66.125) & 66.170 (72.248) & -0.282 & 0.778 \\
    PPG & PPG\_Rate\_Mean & -- & -- & -- & -- & -- & -- & -- & -- & 72.166 (7.225) & 72.364 (7.406) & -0.467 & 0.641 & -0.656 (4.154) & -0.445 (4.153) & -2.013 & {\cellcolor[HTML]{D3D3D3}0.048} \\
     & RSP\_Mean & 0.238 (3.824) & 0.081 (3.782) & 1.396 & 0.164 & 0.214 (3.838) & 0.027 (3.787) & 1.587 & 0.114 & 0.398 (3.635) & 0.249 (3.704) & 1.812 & 0.072 & 1.663 (1.282) & 1.801 (1.370) & -1.542 & 0.128 \\
     & RSP\_Std\_dev & 1.438 (1.323) & 1.549 (1.382) & -1.976 & {\cellcolor[HTML]{D3D3D3}0.049} & 1.505 (1.396) & 1.690 (1.489) & -2.728 & {\cellcolor[HTML]{D3D3D3}0.007} & 1.582 (1.377) & 1.729 (1.489) & -2.720 & {\cellcolor[HTML]{D3D3D3}0.007} & -0.654 (4.213) & -0.279 (4.093) & -2.904 & {\cellcolor[HTML]{D3D3D3}0.005} \\
     & RSP\_Median & 0.224 (3.961) & 0.098 (3.949) & 0.928 & 0.354 & 0.198 (3.989) & 0.088 (3.904) & 0.783 & 0.435 & -0.866 (4.373) & -0.614 (4.313) & -2.231 & {\cellcolor[HTML]{D3D3D3}0.028} & 1.999 (4.018) & 2.248 (3.979) & -3.052 & {\cellcolor[HTML]{D3D3D3}0.003} \\
     & RSP\_Max\_value & 2.364 (3.151) & 2.418 (2.882) & -0.543 & 0.587 & 2.511 (3.114) & 2.605 (2.841) & -0.974 & 0.331 & 1.577 (4.360) & 1.721 (4.368) & -3.251 & {\cellcolor[HTML]{D3D3D3}0.002} & 2.979 (2.549) & 2.903 (2.769) & 0.874 & 0.383 \\
     & RSP\_Min\_value & -2.004 (5.069) & -2.278 (5.132) & 2.084 & {\cellcolor[HTML]{D3D3D3}0.038} & -2.259 (5.237) & -2.657 (5.349) & 2.627 & {\cellcolor[HTML]{D3D3D3}0.009} & -2.534 (5.251) & -2.915 (5.464) & 2.306 & {\cellcolor[HTML]{D3D3D3}0.022} & -3.123 (5.286) & -3.344 (5.516) & 1.073 & 0.285 \\
     & RSP\_Skewness & 0.118 (0.660) & 0.189 (0.730) & -1.634 & 0.103 & 0.121 (0.746) & 0.169 (0.763) & -0.746 & 0.457 & 0.043 (0.941) & 0.010 (1.035) & 0.394 & 0.694 & -0.066 (1.037) & -0.034 (0.976) & -0.375 & 0.708 \\
    \multirow{-7}{*}{RSP} & RSP\_Kurtosis & -0.893 (1.228) & -0.788 (1.814) & -0.934 & 0.351 & -0.672 (1.890) & -0.669 (1.775) & -0.020 & 0.984 & -0.175 (3.098) & -0.033 (3.865) & -0.477 & 0.634 & 0.118 (3.364) & -0.009 (3.122) & 0.461 & 0.646 \\
     & EDA\_Mean & 6.312 (4.961) & 6.346 (5.224) & -2.217 & {\cellcolor[HTML]{D3D3D3}0.029} & 6.618 (5.243) & 6.410 (5.233) & 0.467 & 0.641 & 6.715 (5.513) & 6.426 (5.473) & 4.303 & {\cellcolor[HTML]{D3D3D3}0.000} & 6.699 (5.391) & 6.246 (5.293) & 3.854 & {\cellcolor[HTML]{D3D3D3}0.000} \\
     & EDA\_Std & 0.002 (0.003) & 0.002 (0.006) & -0.119 & 0.905 & 0.011 (0.019) & 0.012 (0.031) & -0.835 & 0.405 & 0.095 (0.127) & 0.091 (0.265) & 0.171 & 0.864 & 0.197 (0.151) & 0.182 (0.448) & 0.402 & 0.688 \\
     & EDA\_Max & 6.315 (4.962) & 6.349 (5.225) & -0.119 & 0.905 & 6.637 (5.248) & 6.430 (5.236) & 0.466 & 0.642 & 6.880 (5.567) & 6.568 (5.550) & 3.694 & {\cellcolor[HTML]{D3D3D3}0.000} & 7.051 (5.490) & 6.519 (5.466) & 3.479 & {\cellcolor[HTML]{D3D3D3}0.001} \\
     & EDA\_Min & 6.310 (4.961) & 6.343 (5.224) & -0.076 & 0.939 & 6.603 (5.234) & 6.390 (5.232) & 0.479 & 0.633 & 6.586 (5.465) & 6.284 (5.404) & 6.036 & {\cellcolor[HTML]{D3D3D3}0.000} & 6.458 (5.342) & 5.966 (5.106) & 8.653 & {\cellcolor[HTML]{D3D3D3}0.000} \\
     & EDA\_Range & 0.006 (0.010) & 0.006 (0.019) & -0.553 & 0.581 & 0.034 (0.061) & 0.040 (0.100) & -0.780 & 0.436 & 0.295 (0.402) & 0.284 (0.820) & 0.179 & 0.858 & 0.592 (0.454) & 0.553 (1.388) & 0.329 & 0.743 \\
     & EDA\_Num\_peaks & 2.496 (1.485) & 2.619 (1.512) & -0.965 & 0.335 & 3.624 (2.782) & 4.392 (2.709) & -2.713 & {\cellcolor[HTML]{D3D3D3}0.008} & 6.245 (5.026) & 7.184 (6.137) & -1.392 & 0.167 & 7.313 (6.318) & 6.597 (6.760) & 0.759 & 0.450 \\
    \multirow{-7}{*}{EDA} & EDA\_Mean\_amplitude & 0.044 (0.102) & 0.025 (0.082) & 2.361 & {\cellcolor[HTML]{D3D3D3}0.019} & 0.070 (0.170) & 0.055 (0.149) & 0.861 & 0.391 & 0.101 (0.261) & 0.093 (0.224) & 0.324 & 0.747 & 0.110 (0.252) & 0.099 (0.165) & 0.487 & 0.628 \\
     & SKT\_Mean & 30.933 (1.096) & 30.933 (1.096) & 1.065 & 0.288 & 30.911 (1.090) & 30.910 (1.089) & 0.793 & 0.428 & 30.917 (1.108) & 30.915 (1.108) & 1.233 & 0.219 & 30.908 (1.095) & 30.906 (1.095) & 1.003 & 0.318 \\
     & SKT\_Std & 0.002 (0.003) & 0.002 (0.002) & 1.499 & 0.135 & 0.003 (0.004) & 0.003 (0.004) & 1.498 & 0.136 & 0.005 (0.005) & 0.004 (0.005) & 1.469 & 0.143 & 0.007 (0.007) & 0.006 (0.007) & 2.200 & {\cellcolor[HTML]{D3D3D3}0.029} \\
     & SKT\_Max & 30.939 (1.094) & 30.937 (1.094) & 1.751 & 0.081 & 30.918 (1.088) & 30.916 (1.087) & 1.535 & 0.126 & 30.926 (1.104) & 30.924 (1.104) & 2.018 & {\cellcolor[HTML]{D3D3D3}0.045} & 30.921 (1.088) & 30.917 (1.090) & 1.919 & 0.057 \\
     & SKT\_Min & 30.928 (1.097) & 30.928 (1.097) & 0.366 & 0.715 & 30.904 (1.091) & 30.904 (1.091) & 0.019 & 0.985 & 30.907 (1.112) & 30.907 (1.112) & 0.273 & 0.785 & 30.895 (1.102) & 30.895 (1.102) & 0.367 & 0.714 \\
     & SKT\_Range & 0.010 (0.009) & 0.009 (0.008) & 1.423 & 0.156 & 0.014 (0.012) & 0.012 (0.011) & 1.469 & 0.143 & 0.019 (0.017) & 0.017 (0.016) & 1.801 & 0.073 & 0.026 (0.023) & 0.023 (0.023) & 1.521 & 0.131 \\
    \multirow{-6}{*}{SKT} & SKT\_Slope & -0.000 (0.000) & -0.000 (0.000) & -1.815 & 0.071 & -0.000 (0.000) & 0.000 (0.000) & -2.629 & {\cellcolor[HTML]{D3D3D3}0.009} & -0.000 (0.000) & -0.000 (0.000) & 1.000 & 0.319 & -0.000 (0.000) & -0.000 (0.000) & -0.332 & 0.740 \\ \hline
    \end{tabular}%
    }
    \begin{tablenotes}
        \fontsize{8pt}{10pt}\selectfont 
        \item[1] The symbol '-' indicates that certain metrics are typically computed over longer time windows and cannot be calculated within the short epoch duration.
        \item[2] The p-values highlighted with gray shading in the table denote statistical significance at the level of $p<0.05$.
    \end{tablenotes}
    \end{threeparttable}
    \end{table}
\end{landscape}

\subsubsection{MER based on Visual Information}
\label{MER-visual}

\begin{table*}[t]
    \centering
    \small
    \renewcommand{\arraystretch}{1.1}
    \caption{Comparative performance of baseline MER methods on the MMME dataset for three-class and seven-class classification}
    \label{tab: three-class and seven-class}
    \resizebox{0.9\textwidth}{!}{%
    \begin{tabular}{clcccccccc}
    \hline
    \toprule[0.5pt]
    \multicolumn{2}{c}{\multirow{2}{*}{\textbf{Methods}}} & \multicolumn{4}{c}{\textbf{Three-Class}} & \multicolumn{4}{c}{\textbf{Seven-Class}} \\ \cmidrule(r){3-6} \cmidrule(r){7-10}
    \multicolumn{2}{c}{} & \textbf{Acc (\%)} & \textbf{F1} & \textbf{UF1} & \textbf{UAR} & \textbf{Acc (\%)} & \textbf{F1} & \textbf{UF1} & \textbf{UAR} \\ \hline
    \multirow{3}{*}{\begin{tabular}[c]{@{}c@{}}Hand-Crafted\\ MER\end{tabular}} & LBP-TOP (2007) \cite{zhao2007dynamic} & 40.79 & 0.3957 & 0.3303 & 0.3443 & 26.56 & 0.2272 & 0.1866 & 0.1741 \\
     & MDMO (2015) \cite{liu2015main} & 44.22 & 0.4534 & 0.4416 & 0.3228 & 29.94 & 0.2634 & 0.2524 & 0.2623 \\
     & Bi-WOOF (2018) \cite{liong2018less} & 50.26 & 0.4823 & 0.4111 & 0.4094 & 34.52 & 0.3309 & 0.2805 & 0.2722 \\ \hline
    \multirow{6}{*}{\begin{tabular}[c]{@{}c@{}}Deep Learning\\ MER\end{tabular}} & RCN-A (2020) \cite{xia2020revealing} & 65.35 & 0.6234 & 0.5635 & 0.5424 & 38.78 & 0.3502 & 0.2981 & 0.2526 \\
     & MERSiamC3D (2021) \cite{zhao2021two} & 70.89 & 0.6827 & 0.6625 & 0.6344 & 40.43 & 0.4101 & 0.3092 & 0.3246 \\
     & FeatRef (2022) \cite{zhou2022feature} & 72.46 & 0.7080 & 0.6339 & 0.6075 & 43.34 & 0.4234 & 0.3543 & 0.3303\\
     & MoExt (2024) \cite{li2024micro} & \underline{74.23} & 0.7254 & 0.6824 & 0.7035 & \underline{46.27} & \underline{0.4541} & \textbf{0.3782} & 0.3475 \\ 
     & HTNet (2024) \cite{wang2024htnet} & \textbf{74.96} & \underline{0.7326} & \textbf{0.7072} & \textbf{0.7171} &  \textbf{47.64} & \textbf{0.4543} & \underline{0.3766} & \textbf{0.3550} \\
     & CSARNet (2025) \cite{zhao2025channel} & 74.05 & \textbf{0.7576} & \underline{0.6958} & \underline{0.7042} & 44.55 & 0.4434 & 0.3582 & \underline{0.3529} \\
     \hline
    \toprule[0.5pt]
    \end{tabular}%
    }
        \begin{tablenotes}
            \fontsize{8pt}{9pt}\selectfont 
            \item[1] \qquad\ $^1$ Three-class classification includes negative, positive, and neutral emotions; seven-class classification comprises happiness, \item[1] \qquad\ \quad\ sadness, surprise, fear, anger, disgust, and contempt.
            \item[2] \qquad\ $^2$ The best results are highlighted in bold, and the second best are underlined.
        \end{tablenotes}
\end{table*}

\begin{table*}[t]
    \centering
    \small
    \renewcommand{\arraystretch}{1.1}
    \caption{Comparative performance of baseline MER methods across five datasets for three-class classification}
    \label{tab: five datasets}
    \resizebox{0.9\textwidth}{!}{%
    \begin{tabular}{lcccccccccc}
    \hline
    \toprule[0.5pt]
    \multirow{2}{*}{\textbf{Methods}} & \multicolumn{2}{c}{\textbf{SMIC}} & \multicolumn{2}{c}{\textbf{CASME II}} & \multicolumn{2}{c}{\textbf{SAMM}} & \multicolumn{2}{c}{\textbf{CAS(ME)$^3$}} & \multicolumn{2}{c}{\textbf{MMME}} \\
    \cmidrule(r){2-3} \cmidrule(r){4-5} \cmidrule(r){6-7} \cmidrule(r){8-9} \cmidrule(r){10-11}
     & \textbf{UF1} & \textbf{UAR} & \textbf{UF1} & \textbf{UAR} & \textbf{UF1} & \textbf{UAR} & \textbf{UF1} & \textbf{UAR} & \textbf{UF1} & \textbf{UAR} \\ \hline
    RCN-A (2020) \cite{xia2020revealing} & 0.6326 & 0.6441 & 0.8512 & 0.8123 & 0.7601 & 0.6715 & 0.3928 & 0.3893 & 0.5635 & 0.5424 \\
    MERSiamC3D (2021) \cite{zhao2021two} & 0.7356 & 0.7598 & 0.8818 & 0.8763 & 0.7475 & 0.7280 & - & - & 0.6625 & 0.6344 \\
    FeatRef (2022) \cite{zhou2022feature} & 0.7011 & 0.7083 & 0.8915 & 0.8873 & 0.7372 & 0.7155 & 0.3493 & 0.3413 & 0.6339 & 0.6075 \\ 
    MoExt (2024) \cite{li2024micro} & 0.7976 & 0.7803 & 0.8992 & 0.8912 & 0.8135 & 0.8100 & 0.5457 & 0.5784 & 0.6824 & 0.7035 \\
    HTNet (2024) \cite{wang2024htnet} & 0.8049 & 0.7905 & 0.9101 & 0.9119 & 0.8131 & 0.8124 & 0.5767 & 0.5415 & 0.7072 & 0.7171 \\
    CSARNet (2025) \cite{zhao2025channel} & 0.7605 & 0.7639 & 0.9254 & 0.9298 & 0.7894 & 0.7924 & - & - & 0.6958 & 0.7042 \\\hline
    \toprule[0.5pt]
    \end{tabular}%
    }
        \begin{tablenotes} 
            \fontsize{8pt}{8pt}\selectfont 
            \item[1] \qquad\ $^1$ The '-' in the table indicates that the information is not provided in the original paper.
        \end{tablenotes}
\end{table*}

\begin{figure*}[t]
    \centering
    \includegraphics[width=0.95\linewidth]{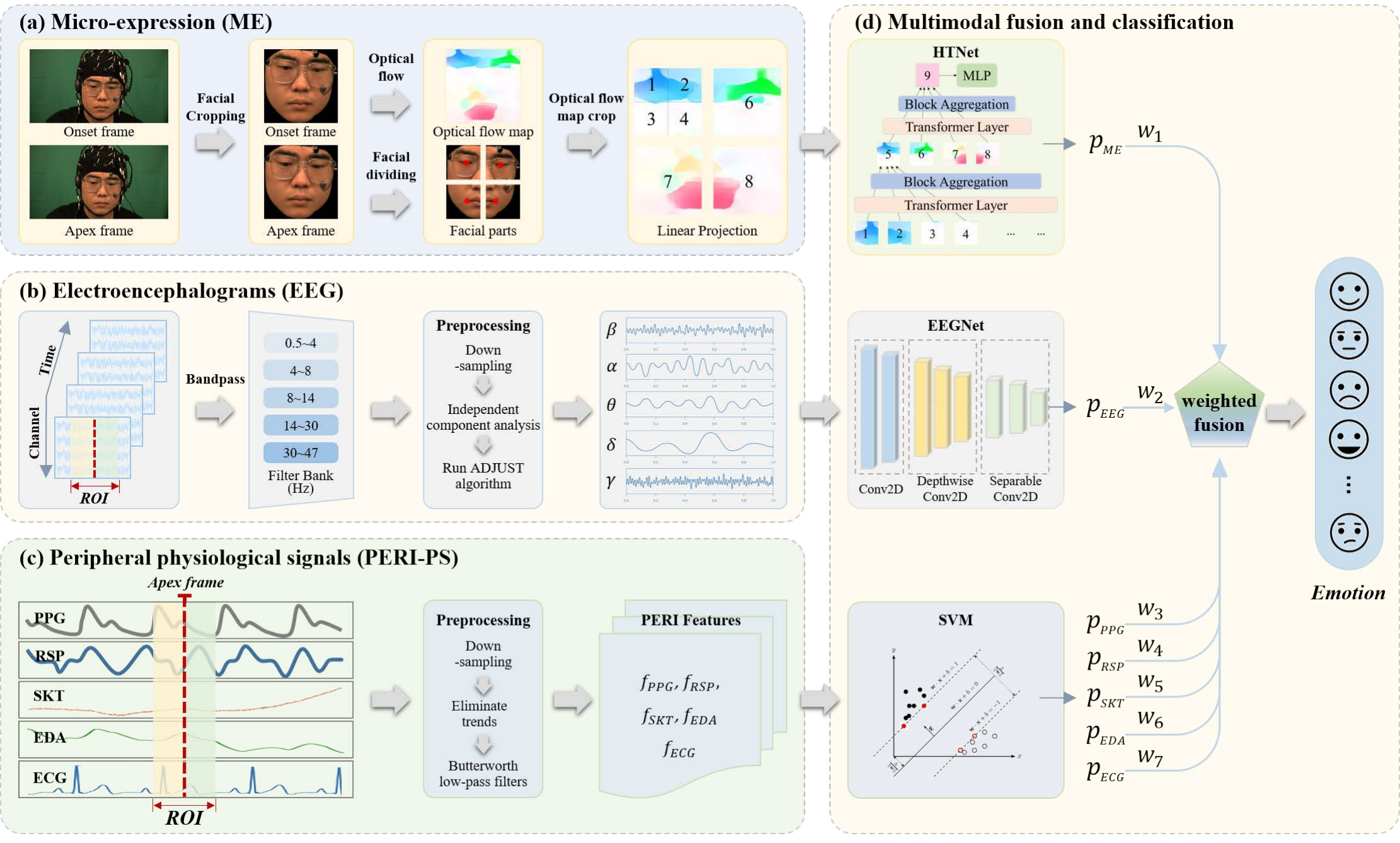}
    \caption{The proposed multimodal MER framework integrating MEs and PS. The framework first performs preprocessing and feature extraction for each modality, followed by decision-level fusion through a weighted voting strategy to combine the outputs of multiple classifiers.}
    \label{fig:Methodology}
\end{figure*}

\textbf{Data Preprocessing.} To effectively reduce the interference of non-facial regions in ME frame sequences during feature analysis, all ME videos were preprocessed prior to the recognition experiments, as illustrated in Fig. \ref{fig:Methodology}(a). First, leveraging the distinct separability between facial skin tones and the green background in the HSV color space, precise segmentation of the facial region was achieved. Subsequently, a face detection algorithm\footnote{\url{https://vision.aliyun.com/facebody}} provided by Alibaba Cloud was employed to identify and crop the facial area in the initial frame. This region was then consistently applied to crop the same area across all subsequent frames in the sequence. This approach is based on the observation that MEs are brief, subtle facial movements with minimal positional changes between consecutive frames, whereas frame-by-frame face detection may introduce inconsistencies and disrupt the temporal continuity of MEs. Finally, all cropped facial regions were resized to a uniform resolution of 128 $\times$ 128 pixels to ensure consistency in feature extraction and to enhance computational efficiency.

\textbf{MER performance.} In this section, we adopted two categories of benchmark methods: handcrafted feature-based approaches and deep learning-based approaches. The handcrafted methods include LBP-TOP \cite{wei2022micro}, MDMO \cite{lu2024learning}, and Bi-WOOF \cite{liong2018less}. LBP-TOP extends local binary patterns to three dimensions by integrating texture features from X-Y, X-T, and Y-T planes. MDMO employs optical flow across 36 facial regions of interest to detect micro-movements, while Bi-WOOF uses dual-weighted optical flow and strain magnitudes to enhance apex frame discrimination. The evaluated deep learning methods comprise RCN-A \cite{xia2020revealing}, MERSiamC3D \cite{zhao2021two}, FeatRef \cite{zhou2022feature}, MoExt \cite{li2024micro}, HTNet \cite{wang2024htnet}, and CSARNet \cite{zhao2025channel}. RCN-A adapts a recurrent convolutional network with three parameter-free modules to improve representation across multiple perspectives, maintaining a shallow architecture suitable for lower-resolution data. MERSiamC3D introduces a two-stage learning framework, involving prior and target learning, based on a Siamese 3D convolutional neural network for MER. FeatRef proposes a feature refinement approach with an expression proposal module incorporating an attention mechanism and a classification branch. MoExt develops a motion extraction strategy, pretraining the feature separator and motion extractor with contrastive loss to capture representative motion features. HTNet leverages hierarchical transformers to identify key facial movement regions through local temporal and global semantic representations. CSARNet implements a local feature augmentation strategy to enhance the local feature representations of motion flow images. Additionally, it incorporates a lightweight backbone network designed to reduce model complexity and computational time while accurately extracting discriminative ME information across channel and spatial dimensions, thereby facilitating effective emotion recognition. For fairness, all hyperparameters—including feature map size, batch size, and learning rate—are kept consistent with the default configurations reported in the original papers. During the evaluation, we adopted several commonly used metrics from the MER task in MEGC2019 \cite{see2019megc}, including Accuracy (Acc), F1-score, Unweighted F1 Score (UF1), and Unweighted Average Recall (UAR), along with Leave-One-Subject-Out Cross-Validation (LOSO-CV) protocol. 

Table \ref{tab: three-class and seven-class} presents the comparative performance of baseline MER methods on the MMME dataset for three-class and seven-class classification, demonstrating that the samples in the MMME dataset exhibit good separability and discriminative capability. For instance, the Bi-WOOF model achieved an accuracy of 50.26\% in the three-class classification task and 34.52\% in the seven-class task, while the HTNet model attained 74.96\% and 47.64\% accuracy, respectively. The deep learning methods outperformed traditional handcrafted feature-based approaches, owing to their superior ability to extract discriminative spatiotemporal features and model subtle facial movement patterns through nonlinear representations. Among these deep learning models, HTNet achieves the highest performance, owing to its hierarchical architecture that facilitates multi-scale feature extraction, significantly improving the accuracy of MER. Then, we conducted comprehensive comparisons of these deep learning methods on five benchmark ME datasets (SAMM, CASME II, SMIC, CAS(ME)$^{3}$, and our proposed MMME) for three-class classification, as detailed in Table \ref{tab: five datasets}. The experimental results demonstrate that these models achieved competitive recognition performance on the MMME dataset, with UF1 and UAR metrics generally outperforming those on the  CAS(ME)$^{3}$ dataset.

\subsubsection{MER based on Multimodal Fusion}

Recent studies have increasingly highlighted the importance of incorporating multimodal data in ME analysis, with the dual objectives of leveraging more comprehensive emotional representations and overcoming the limitations of unimodal approaches. In this section, we proposed a multimodal fusion-based MER framework to validate the effectiveness of the MMME dataset in multimodal research. The pipeline is outlined as follows:

\textbf{Data Preprocessing.} For MEs, the key frames were processed following the data preprocessing procedure described in Section \ref{MER-visual}. The EEG signal was first passed through a filter bank, decomposing it into five frequency bands: 0.5–4 Hz ($\delta$), 4–8 Hz ($\theta$), 8–14 Hz ($\alpha$), 14–30 Hz ($\beta$), and 30–47 Hz ($\gamma$). Subsequent preprocessing involved down-sampling to reduce data complexity, followed by independent component analysis (ICA) to isolate distinct signal components. The Adjust algorithm was then applied to remove artifacts and noise. The resulting output yielded cleaned waveforms corresponding to the $\beta$, $\alpha$, $\theta$, $\delta$, and $\gamma$ bands, facilitating further analysis of brain activity patterns. The PERI-PS analyzed in this section include PPG, ECG, RSP, SKT, and EDA. Initially recorded at a sampling rate of 1000 Hz, these signals were downsampled to 200 Hz to reduce processing time. Low-frequency drifts over time were removed to eliminate trends in the signals. To reduce noise, Butterworth low-pass filters with different cutoff frequencies were applied based on the characteristics of each signal. For the ECG signal, a Butterworth low-pass filter with a cutoff frequency of 40 Hz was used to remove high-frequency noise. As the EDA signal reflects slow changes in skin conductance, a lower cutoff frequency of 5 Hz was applied to retain these slow changes while eliminating faster fluctuations or noise. For RSP, a low-pass cutoff frequency of 1 Hz was used in the Butterworth filter.

\textbf{Feature Extraction.} To effectively capture the motion characteristics of MEs, this study utilized the FlowNet 2.0 algorithm\footnote{\url{https://github.com/NVIDIA/flownet2-pytorch}} to extract optical flow features, focusing on subtle movements in MEs. A schematic diagram of this process is depicted in Fig. \ref{fig:Methodology}(a). The optical flow feature map is formulated as follows:

    \begin{equation}
    \small
        V=(u(x, y), v(x, y)) \mid x=1,2, \ldots ., X, y=1,2, \ldots, Y,
        \label{formulation: optical flow}
    \end{equation}
where $X$ and $Y$ represent the width ($W$) and height ($H$) of the image, respectively, and $u(x, y)$ and $v(x, y)$ denote the horizontal and vertical component of optical flow feature map $V$ ($V=\left[V_{x}, V_{y}\right]$), with $V \in R^{W \times H \times 2}$. Additionally, the optical strain provides insights into the degree of facial displacement, thereby offering valuable information about the subtle movements during MEs. The optical strain is derived from the gradient of the optical flow field and is expressed as:

\begin{equation}
    V_{z}=\sqrt{\frac{\partial V_{x}^{2}}{\partial x}+{\frac{\partial V_{y}^{2}}{\partial y}}+\frac{1}{2}\left({\frac{\partial V_{x}^{2}}{\partial y}}+{\frac{\partial V_{y}^{2}}{\partial x}}\right)},
\end{equation}
where ${\frac{\partial V_{x}}{\partial x}}^{2}$, ${\frac{\partial V_{y}}{\partial y}}^{2}$, ${\frac{\partial V_{x}}{\partial y}}^{2}$ and ${\frac{\partial V_{y}}{\partial x}}^{2}$ are the partial first-order derivatives of $V$. Finally, three-dimension optical flow feature maps are represented as $V_{m}=\left[V_{x}, V_{y}, V_{z}\right]$, with $V_{m} \in R^{W \times H \times 3}$.

\begin{table}[t]
    \centering
    \caption{List of extracted expert-defined and statistical features from the recorded PERI-PS}
    \label{tab:features}
    \Huge
    \resizebox{\columnwidth}{!}{%
    \begin{tabular}{cclc}
    \hline
    \toprule[0.5pt]
    \textbf{Modality} & \textbf{Features} & \multicolumn{1}{c}{\textbf{Description}} & \textbf{Reference} \\ \hline
    PPG & $PR$ & \makecell[l]{Pulse rate} & \cite{sabour2019emotional} \\
    \multirow{10}{*}{ECG} & $HR$ & Mean heart rate & \cite{lane2009neural} \\  
     & $SDSD$ & \makecell[l]{Standard deviation of successive NN \\ intervals} & \cite{agrafioti2011ecg, pham2021heart} \\ 
     & $SDNN$ & Standard deviation of NN intervals & \cite{agrafioti2011ecg, pham2021heart} \\ 
     & $RMSDD$ & \makecell[l]{Root mean square of successive \\ differences} & \cite{agrafioti2011ecg, pham2021heart} \\ 
     & $LF, HF$ & \begin{tabular}[c]{@{}l@{}}\makecell[l]{Low and high-frequency component \\ in the range of 0.05-0.15 Hz and \\ 0.15-0.4 Hz of the Welch spectrum}\end{tabular} & \cite{agrafioti2011ecg, pham2021heart} \\ 
     & $LF_{n}, HF_{n}$ & Normalized HF and LF spectrum & \cite{agrafioti2011ecg, pham2021heart} \\ 
     & $LF/HF$ & Ratio of the LF and HF component & \cite{agrafioti2011ecg, pham2021heart} \\ 
     & $SD1/SD2$ & \begin{tabular}[c]{@{}l@{}}\makecell[l]{Ratio of SD1 and SD2 as standard \\ deviation along the identity lines of \\ a Poincaré plot}\end{tabular} & \cite{agrafioti2011ecg, pham2021heart} \\ 
     & $PSS$ & Percentage of short segments & \cite{costa2017heart, hayano2020impact} \\  
     & $PIP$ & \makecell[l]{Percentage of inflection points in NN \\ intervals} & \cite{costa2017heart, hayano2020impact} \\
    \multirow{2}{*}{RSP} & $\mu_{BR}, \sigma_{BR}$ & \makecell[l]{Mean and standard deviation of \\ breathing rate} & \cite{grassmann2016respiratory} \\ 
     & $\mu_{[E-I]}$ & \makecell[l]{Mean of the ratio of inhalation and \\ exhalation amplitude values} & \cite{grassmann2016respiratory} \\
    \multirow{2}{*}{SKT} & $\mu_{T}, \sigma_{T}$ & \makecell[l]{Mean and standard deviation of the \\ skin temperature} & \cite{kim2004emotion, lin2023review} \\ 
     & $min_{T}, max_{T}$ & \makecell[l]{Minimal and maximal of the \\ temperature} & \cite{kim2004emotion, lin2023review} \\
    \multirow{5}{*}{EDA} & $\sum_{t}^{W} S C L^{\prime}(t)$ & \makecell[l]{Change of the skin conductance level} & \cite{giannakakis2019review, yu2020systematic, rahma2022electrodermal} \\ 
     & $\#_{SCR}^{Peaks} / W$ & \makecell[l]{Number of peaks as measure of the \\ phasic component} & \cite{giannakakis2019review, yu2020systematic, rahma2022electrodermal} \\  
     & $\mu_{SCR}^{Amplitude}$ & Mean amplitude values of SCR peaks & \cite{giannakakis2019review, yu2020systematic, rahma2022electrodermal} \\  
     & $\mu_{SCR}^{Rise}$ & Mean rise time of SCR peaks & \cite{giannakakis2019review, yu2020systematic, rahma2022electrodermal} \\  
     & $\mu_{SCR}^{Recovery}$ & \makecell[l]{Mean recovery time to 50 percent of \\ the maximal peak amplitude} & \cite{giannakakis2019review, yu2020systematic, rahma2022electrodermal} \\ 
     \toprule[0.5pt]
    \end{tabular}%
    }
\end{table}

Inspired by Saffaryazdi \cite{saffaryazdi2022using}, we consider the apex frame as the point of peak emotional intensity in each trial. A fixed-length time window centered on the apex frame is designated as the region of interest (ROI), and only the physiological data within this window are used for analysis. When selecting the ROI window size, we observed that an excessively large window may cause overlap between ROIs of adjacent MEs, while a window that is too small may result in the loss of important features. Based on these observations, we set the window size to 10 s. Based on the concordance analysis between MEs and PERI-PS in Section \ref{subsection-concordance}, relevant literatures, and medical expert knowledge, this study extracted a series of PERI-PS features significantly associated with emotions, as shown in Table \ref{tab:features}. PPG was used to assess the mechanical activity of the heart, from which pulse rate (PR) was derived to reflect variations in cardiac rhythm. ECG, by contrast, recorded the heart's electrical activity. In addition to heart rate (HR), we extracted HRV features, including the standard deviation of successive NN intervals (SDSD), the standard deviation of NN intervals (SDNN), and the root mean square of successive NN interval differences (RMSSD), all of which are closely linked to emotional fluctuations. RSP patterns were obtained by detecting the extrema of breathing cycles, with features such as amplitude differences, mean, and standard deviation of the respiration rate. For SKT, we extracted statistical features including the mean, standard deviation, maximum, and minimum values. EDA reflects sympathetic nervous system arousal, typically indicated by marked changes during emotional stress. We analyzed both the skin conductance level (SCL), representing slower changes, and the skin conductance response (SCR), reflecting rapid fluctuations. Finally, all signals underwent normalization. The preprocessing and feature extraction procedures for the PS were implemented using NeuroKit2\footnote{\url{https://neuropsychology.github.io/NeuroKit/index.html}}, an open-source neurophysiological processing toolbox. The resulting emotion-related features can be mathematically defined as follows:
    \begin{equation}
    f_{PERI-PS}=[f_{PPG}, f_{ECG}, f_{RSP}, f_{SKT}, f_{EDA}].   
    \end{equation}

\textbf{Multimodal Fusion and Classification.} To improve MER performance, we implement a multi-classifier voting framework that capitalizes on the complementary discriminative capabilities of diverse classifiers. The system integrates seven modality-specific sub-classifiers: the ME classifier utilizes HTNet \cite{wang2024htnet}, the EEG classifier employs EEGNet \cite{lawhern2018eegnet}, and the five PERI-PS classifiers utilize SVMs. HTNet includes two major components: a transformer layer that leverages the local temporal features and an aggregation layer that extracts local and global semantical facial features, achieving optimal performance on the MMME dataset, as shown in Table \ref{tab: three-class and seven-class}. EEGNet is chosen due to its compact, efficient convolutional neural network architecture, tailored for processing complex EEG signal. SVMs are adopted for PERI-PS classification due to the lower-dimensional and less complex nature of these signals compared to ME or EEG data, as SVMs excel at identifying optimal decision boundaries for PS features. Following individual emotion classification by each sub-classifier, a weighted voting scheme synthesizes the final prediction. A novel feedback-driven weighting mechanism was developed to optimize decision fusion. This approach dynamically adjusts voting weights by incorporating both the inherent reliability of each modality, and the real-time classification accuracy of PS-based emotion recognition. The methodological details of this adaptive weighting strategy are illustrated in Fig. \ref{fig:Methodology}(c), comprising the following key steps:

\textbf{Step 1:} Calculate the weighted matrix $W_{i}$ for each sub-classifier based on the classification accuracy of its respective modality. Each sub-classifier, corresponding to a single modality, is independently trained and tested to determine the recognition accuracy for each emotional state:

    \begin{equation}
        \vec{P}_{i}=\left(p_{i1}, \ldots, p_{ij}\right)^{T}, i \in\{1,2, \ldots, 7\},
    \end{equation}
where $p_{ij} \in [0,1]$ represents the predicted probability of the $j$-th emotional state by the $i$-th sub-classifier, and $j \in \{1,2, \ldots, m\}$. Using a feedback-based principle, the weighted matrix for each modality is determined as follows:

\begin{equation}
    W_{i}=\left[\begin{array}{ccc}
    p_{i 1} & \cdots & 0 \\
    \vdots & \ddots & \vdots \\
    0 & \cdots & p_{ij}
    \end{array}\right], i \in\{1,2, \ldots, 7\},
\end{equation}
where $W_{i}$ is the weighted matrix of the $i$-th modality.

\textbf{Step 2:} Let $\vec{C}_{i}=\left(c_{i1}, \ldots, c_{ij}\right)^{T} \ (1 \leq i \leq 7)$ represent the recognition result of each sub-classifier, where $\left|\vec{C}_{i}\right|=1$ and $c_{i j} \in\{0,1\}$.
The weighted matrices and recognition results of each sub-classifier are then linearly fused as follows:

\begin{equation}
    \resizebox{.9\hsize}{!}{$\vec{C}=\sum_{i=1}^{7} {W}_{i}\vec{C}_{i}=\sum_{i=1}^{7}\left[\begin{array}{ccc}
    p_{i1} & \cdots & 0 \\
    \vdots & \ddots & \vdots \\
    0 & \cdots & p_{ij}
    \end{array}\right]\left[\begin{array}{c}
    c_{i1} \\
    \vdots \\
    c_{ij}
    \end{array}\right]=\left[\begin{array}{c}
    \sum_{i=1}^{7} c_{i1} p_{i1} \\
    \vdots \\
    \sum_{i=1}^{7} c_{ij} p_{ij}
    \end{array}\right]$.}
\end{equation}

\textbf{Step 3:} Based on the maximum score rule, the emotional state with the highest score, denoted as class $k$ is selected as the final recognition result:

\begin{equation}
    MAX_{j=1}^{m}\left\{\sum_{i=1}^{7} c_{ij} p_{ij}\right\}=\sum_{i=1}^{7} c_{ik} p_{ik}.
\end{equation}

Through this approach, this study effectively integrated the recognition results from multiple sub-classifiers, thereby enhancing the reliability of the final classification. 

    \begin{figure}[b]
        \centering
        \includegraphics[width=0.95\linewidth]{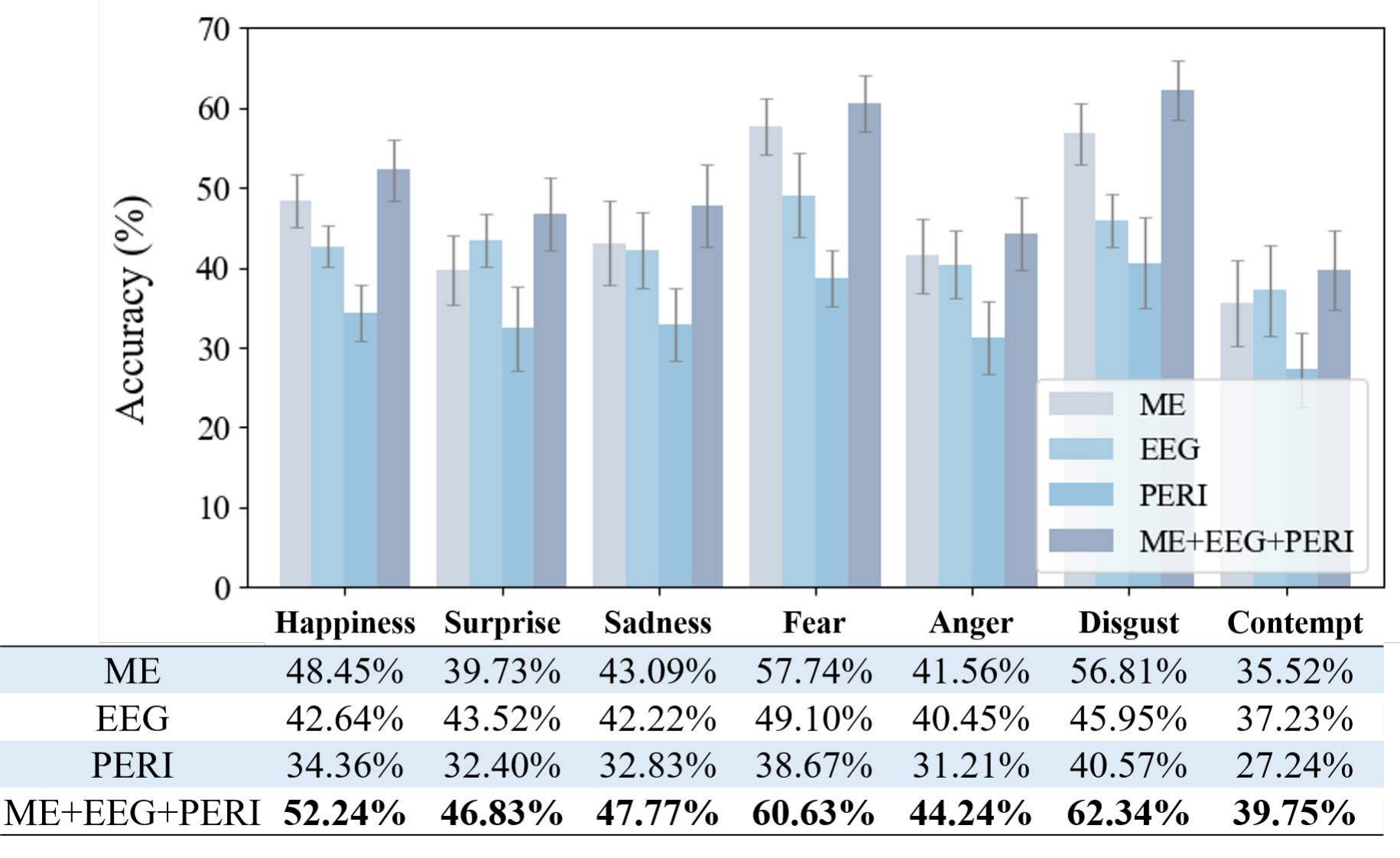}
        \caption{A comparative analysis of seven basic emotion recognition accuracies across different modalities, highlighting the superiority of multimodal fusion.}
        \label{fig:Discrete emotion accuracy}
    \end{figure}

Fig. \ref{fig:Discrete emotion accuracy} presents a comparative evaluation of the recognition accuracy for seven basic emotions across various modalities, encompassing unimodal methods (ME, EEG, and PERI) and a multimodal fusion approach (ME+EEG+PERI). The experimental findings reveal that ME generally achieve superior classification accuracy for the majority of emotions, surpassing unimodal methods reliant on EEG or PERI. For instance, ME achieve the highest recognition rates for fear (57.74\%±3.46), disgust (56.81\%±3.78), and happiness (48.45\%±3.34), potentially due to the more pronounced facial muscle movement features associated with these emotions, which offer greater discriminability compared to PS. However, for certain emotions such as surprise and contempt, the ME based method is less effective than EEG, possibly due to the MEs for these emotions are subtler and harder to capture accurately through facial movements. Although PERI exhibit a relatively lower overall recognition rate, they provide supplementary value in identifying disgust (40.57\%±5.67), highlighting the unique complementary strengths of different modal signals in emotion recognition. By integrating the complementary information from ME, EEG, and PERI, the multimodal fusion method (ME+EEG+PERI) significantly enhances overall emotion recognition performance. For instance, the recognition accuracy for happiness increases from 48.45\% in the ME modality to 52.24\% in the fusion model, for surprise from 39.73\% to 46.83\%, and for sadness from 43.09\% to 47.77\%. These findings substantiate the synergistic effect of visual and PS within the MMME dataset. Additionally, Tables \ref{tab:seven-class emotion classification} and \ref{tab: three-class emotion classification} provide detailed performance metrics, including Acc (\%), F1 score, UF1, and UAR, for different modalities across three-class and seven-class emotion recognition tasks, demonstrating that the fused modality consistently outperforms individual modalities. This further underscores the significant advantages of the multimodal fusion strategy in enhancing emotion recognition efficacy.

\begin{table}[t]
    \Huge
    \centering
    \caption{Seven-class emotion classification performance across different modality combinations}
    \label{tab:seven-class emotion classification}
    \begin{threeparttable}[b]
        \resizebox{\columnwidth}{!}{%
            \begin{tabular}{lcccc}
            \toprule[1pt]
            \textbf{Modality Type} & \textbf{\begin{tabular}[c]{@{}c@{}}Acc(\%)\\ (mean±std)\end{tabular}} & \textbf{\begin{tabular}[c]{@{}c@{}}F1\\ (mean±std)\end{tabular}} & \textbf{\begin{tabular}[c]{@{}c@{}}UF1\\ (mean±std)\end{tabular}} & \textbf{\begin{tabular}[c]{@{}c@{}}UAR\\ (mean±std)\end{tabular}} \\
            \midrule
                ME & 47.64±3.18 & 0.4543±0.0245 & 0.3766±0.0333 & 0.3550±0.0278 \\
                EEG & 42.19±2.47 & 0.4089±0.0311 & 0.3447±0.0256 & 0.3312±0.0305 \\
                PERI & 33.45±3.23 & 0.3256±0.0322 & 0.2998±0.0277 & 0.2776±0.0286 \\
                ME+EEG+PERI & \textbf{49.38±3.22} & \textbf{0.4894±0.0473} & \textbf{0.4021±0.0384} & \textbf{0.3887±0.0423} \\ \hline
            \toprule[0.5pt]
            \end{tabular}%
            }
        \begin{tablenotes}[flushleft]
            \setlength{\leftskip}{0pt} 
            \fontsize{8pt}{9pt}\selectfont
            \item[1] Bold font indicates the best-performing results.
        \end{tablenotes}
    \end{threeparttable}
\end{table}

\begin{table}[t]
    \Huge
    \centering
    \caption{Three-class emotion classification performance across different modality combinations}
    \label{tab: three-class emotion classification}
    \begin{threeparttable}[b]
        \resizebox{\columnwidth}{!}{%
            \begin{tabular}{lcccc}
            \toprule[1pt]
            \textbf{Modality Type} & \textbf{\begin{tabular}[c]{@{}c@{}}Acc(\%)\\ (mean±std)\end{tabular}} & \textbf{\begin{tabular}[c]{@{}c@{}}F1\\ (mean±std)\end{tabular}} & \textbf{\begin{tabular}[c]{@{}c@{}}UF1\\ (mean±std)\end{tabular}} & \textbf{\begin{tabular}[c]{@{}c@{}}UAR\\ (mean±std)\end{tabular}} \\
            \midrule
                ME & 74.96±4.78 & 0.7326±0.0636 & 0.7072±0.0439 & 0.7171±0.0588 \\
                EEG & 69.12±5.66 & 0.6628±0.0572 & 0.5834±0.0480 & 0.5522±0.0452 \\
                PERI & 52.45±3.29 & 0.5034±0.0452 & 0.4205±0.0438 & 0.4137±0.0578\\
                ME+EEG+PERI & \textbf{77.46±5.66} & \textbf{0.7525±0.0470} & \textbf{0.7343±0.0673} & \textbf{0.7455±0.0479} \\ \hline
            \toprule[0.5pt]
            \end{tabular}%
            }
        \begin{tablenotes}[flushleft]
            \setlength{\leftskip}{0pt} 
            \fontsize{8pt}{9pt}\selectfont
            \item[1] Bold font indicates the best-performing results.
        \end{tablenotes}
    \end{threeparttable}
\end{table}

\begin{figure}[t]
    \centering
    \includegraphics[width=0.9\linewidth]{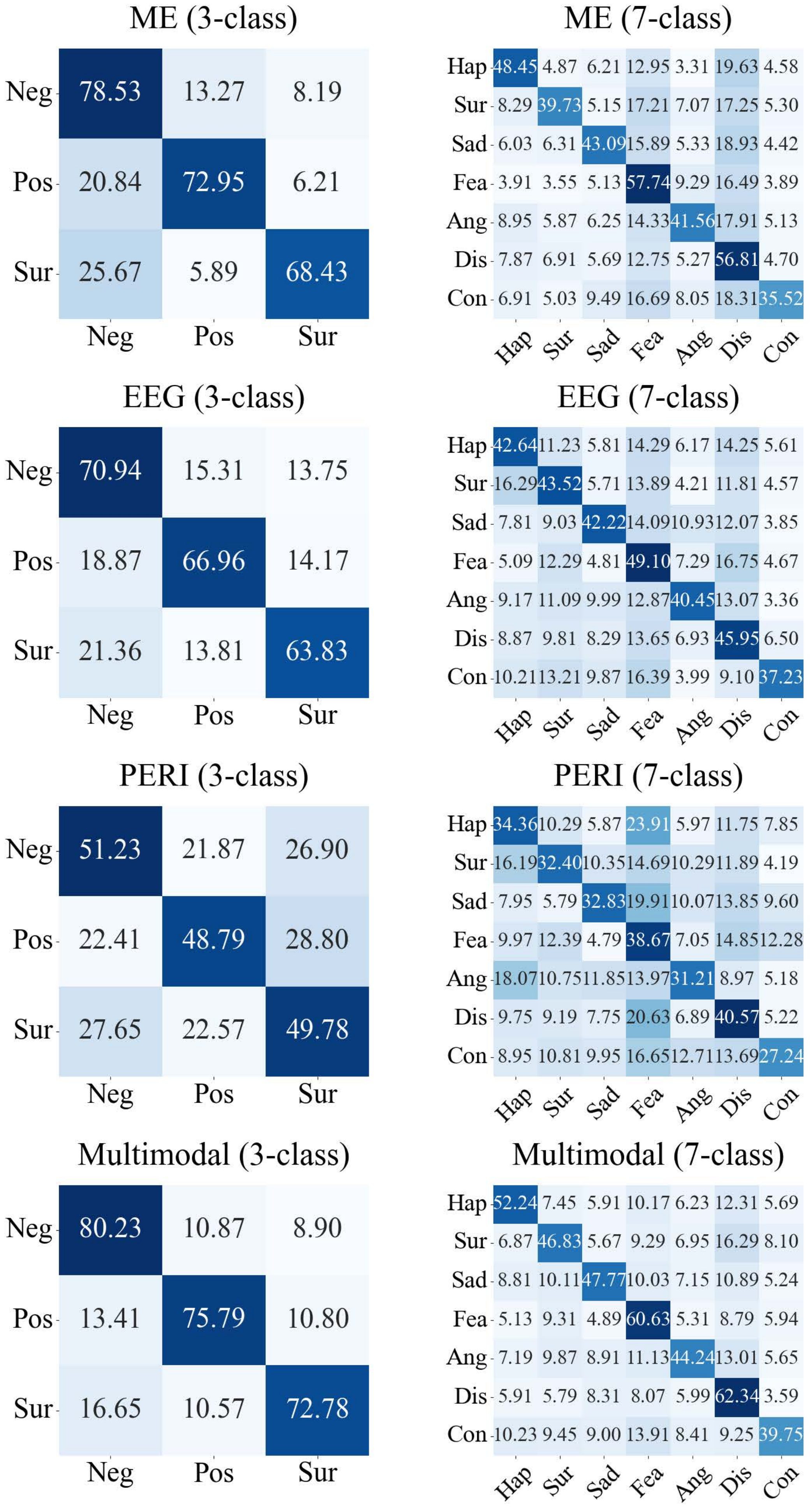}
    \caption{Confusion matrices for emotion classification (three-class and seven-class) across various modalities.}
    \label{figs: Confusion_matrices}
\end{figure}

The confusion matrix in Fig. \ref{figs: Confusion_matrices} clearly illustrates the relationship between the model's predicted results and the true labels, offering a detailed analysis of the model's classification performance. The diagonal elements of the matrix represent the proportion of correctly classified samples, with darker colors indicating higher accuracy. It can be observed that in the three-class task, the classification performance based on ME outperforms EEG, while PERI yields the lowest accuracy. The multimodal fusion approach performs best, with a significantly reduced misclassification rate and a relatively balanced accuracy distribution across classes. In contrast, the seven-class task exhibits a pronounced class imbalance issue. The confusion matrix reveals that the classification accuracy for ``Fear,'' ``Disgust,'' and ``Surprise'' is relatively high, likely due to these emotions being easier to elicit, with more distinct facial expressions and physiological responses. Conversely, the accuracy for ``Anger'' and ``Contempt'' is lower, possibly because these MEs are harder to induce, resulting in fewer samples and consequently poorer model performance on these classes. The multimodal fusion method demonstrates significant advantages in emotion recognition, particularly in fine-grained seven-class tasks, effectively improving classification accuracy. However, certain emotions (e.g., ``Anger'' and ``Contempt'') still exhibit high misclassification rates, suggesting that improvements to the MMME dataset should focus on increasing sample size and enhancing class balance.

\begin{figure}[t]
    \centering
    \includegraphics[width=0.95\linewidth]{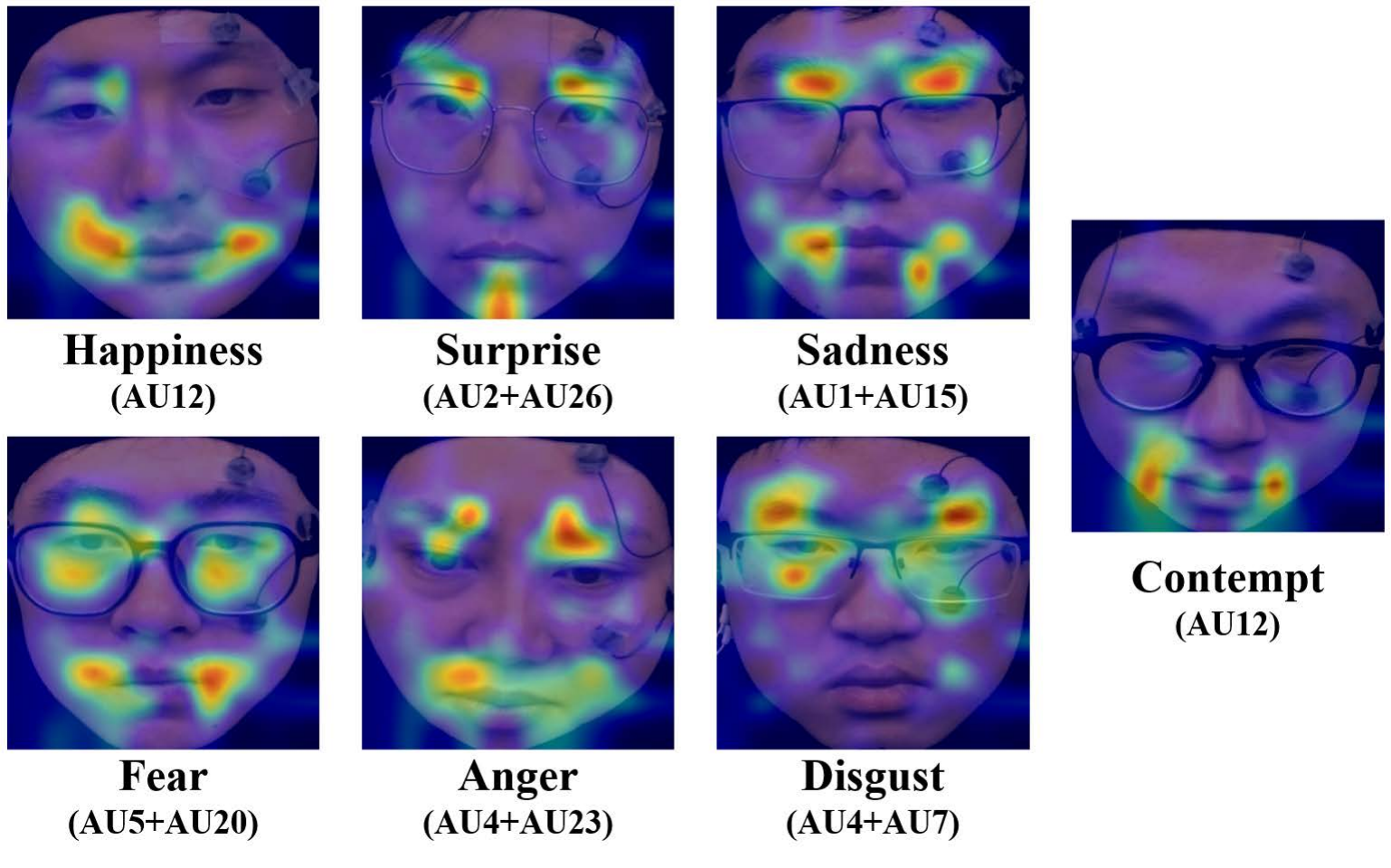}
    \caption{Visual Explanations of ME via Gradient-Based Localization.}
    \label{fig:gradcam}
\end{figure}

\textbf{Visual Analysis.} To gain deeper insights into the learned features of MEs, we utilize Grad-CAM to visualize activation heatmaps, as shown in Fig. \ref{fig:gradcam}. This visualization analysis demonstrates the HTNet model's ability to focus on different regions and visual features in facial images. A sample from each of the seven emotional categories is selected, and Grad-CAM is applied to the output of the model's final convolutional layer. The generated heatmap is then overlaid on the original sample image. The highlighted regions correspond to areas crucial for recognizing subtle MEs, such as the eyebrows and corners of the mouth. For instance, in the ``Happiness" sample, the Grad-CAM heatmap highlights the zygomaticus major muscle, corresponding to AU12, with the action descriptor ``Lip corner puller.'' In the ``Surprise'' sample, the highlighted regions include the frontalis (pars lateralis) and masseter muscles, corresponding to AU2 (``Outer brow raiser'') and AU26 (``Jaw drop''), respectively. In the ``Sadness'' sample, the highlighted region corresponds to the frontalis (pars medialis), associated with AU1 (``Inner brow raiser''). For the ``Contempt'' sample, the zygomaticus major muscle is highlighted, corresponding to AU12 (``Lip corner puller''). These results indicate that the visual feature learning model employed in this study is capable of focusing on key visual regions that are closely related to the target task during the decision-making process. This finding further highlights the pivotal role of visual features in multimodal fusion.

\subsection{ME Spotting Validation}

\begin{figure}[b]
    \centering
    \includegraphics[width=0.95\linewidth]{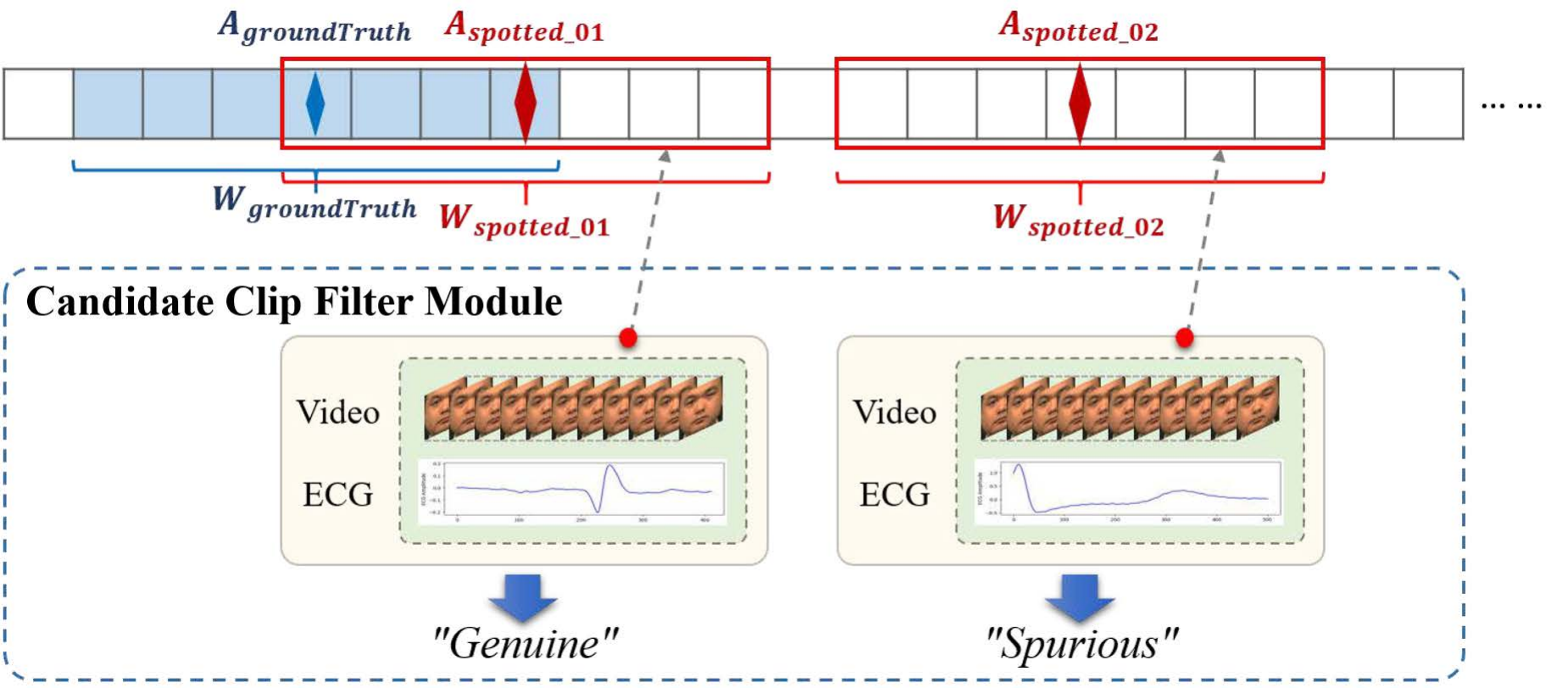}
    \caption{Schematic diagram of the ME spotting pipeline. The blue-filled regions denote the ground-truth ME from manual annotation, while the red-bordered boxes represent the automatically detected results. We propose a Candidate Clip Filter (CCF) module to enhance the accuracy of ME spotting.}
    \label{fig: ME+ECG}
\end{figure}

To further validate the effectiveness and practical value of the proposed MMME dataset, this section investigates ME spotting tasks based on the dataset, encompassing both unimodal (visual-only) and multimodal (visual-ECG fusion) approaches. This serves as a preliminary exploration toward advancing ME spotting from unimodal to multimodal frameworks. Fig. \ref{fig: ME+ECG} illustrates the ME spotting pipeline, which incorporates a Candidate Clip Filter (CCF) module to improve the detection results.
\label{subsection-spotting}

\subsubsection{ME Spotting based on Visual Information}

ME spotting aims to locate the onset, apex, and offset frames of MEs within continuous video streams, and holds significant potential for real-world applications. Considering the characteristics of the dataset, we design two types of detection tasks: a unimodal ME spotting task based solely on video, and a multimodal ME spotting task that incorporates auxiliary ECG signals. These tasks not only facilitate the evaluation of the dataset's quality in terms of spatiotemporal annotations and multimodal synchronization but also enable the exploration of the complementary roles of video and PS in ME spotting, thereby laying a new foundation for the advancement of multimodal ME spotting technologies.

\textbf{Evaluation Metrics.} In ME spotting task, Intersection over Union (IoU) serves as a widely adopted evaluation metric that treats ME spotting as an object detection problem. IoU quantifies detection performance by calculating the degree of overlap between the temporal interval of an ME sample and the interval identified by the detection algorithm. Based on the computed IoU value, a detected interval is classified as either a true positive (TP) or a false positive (FP). Specifically, for a detected interval $W_{spotted}$, if there exists a ground-truth ME interval $W_{groundTruth}$ that satisfies the condition defined in Equation \ref{con: IoU}, the predicted interval is judged as a TP:

    \begin{equation}
    \label{con: IoU}
        \begin{matrix}
        IoU = \dfrac{W_{\text{spotted}} \cap W_{\text{groundTruth}}}{W_{\text{spotted}} \cup W_{\text{groundTruth}}} \\[15pt]  
        \text{True Positive: } IoU \geq \varepsilon \\[10pt]  
        \text{False Positive: } IoU < \varepsilon
        \end{matrix},
    \end{equation}
following most existing studies, we set the value of $\varepsilon$ to 0.5.

\textbf{Experimental Results.} Table \ref{tab: ME spotting} presents a comparative performance of different ME spotting algorithms across four datasets, which not only evaluates the quality and annotation accuracy of MMME but also establishes performance benchmarks for this task. The algorithms used include MESNet \cite{wang2021mesnet}, ABPN \cite{leng2022abpn}, SL-Swin \cite{he2023sl}, DMA-LBP \cite{liu2024duration} and MSOF \cite{yang2025msof}. MESNet utilizes 2D convolutions to extract spatial features and 1D convolutions to capture temporal dynamics, enabling video clip classification as MEs or non-MEs while regressing their temporal boundaries. ABPN computes frame-level auxiliary probabilities to identify apex or boundary frames, leveraging these probabilities to generate precise expression proposals for localization. SL-Swin employs a Transformer-based approach, integrating Shifted Patch Tokenization and Locality Self-Attention into the Swin Transformer backbone, to predict frame probabilities within expression intervals for effective ME spotting. DMA-LBP applies multiple sliding windows of varying scales and patterns to create weak detectors tailored to MEs of specific durations and transition modes, aggregating their outputs through a voting-based module. MSOF incorporates low-pass filtering and empirical mode decomposition to eliminate high-frequency noise, using non-maximum suppression to delineate final expression intervals for accurate ME spotting. Experimental results demonstrate that the MMME dataset effectively distinguishes the performance of various algorithms. For instance, MSOF achieves a significantly higher F1-score (0.3747) compared to MESNet (0.0511), a level of differentiation consistent with performance on public datasets, confirming the suitability of the MMME dataset for evaluating ME spotting algorithms. Furthermore, the dataset exhibits robust effectiveness in ME spotting, performing comparably to established public datasets such as SAMM, CASME II, and CAS(ME)$^3$. These findings underscore the MMME’s high consistency and discriminative power in algorithm performance evaluation, making it well-suited for testing and validating ME spotting algorithms.

\begin{table*}[t]
    \centering
    \caption{Comparative performance of different ME spotting algorithms across four datasets}
    \Huge
    \renewcommand{\arraystretch}{1.1}
    \label{tab: ME spotting}
    \resizebox{0.9\textwidth}{!}{%
    \begin{tabular}{lcccccccccccc}
    \hline
    \toprule[0.5pt]
    \multirow{2}{*}{\textbf{Method}} & \multicolumn{3}{c}{\textbf{SAMM}} & \multicolumn{3}{c}{\textbf{CASME II}} & \multicolumn{3}{c}{\textbf{CAS(ME)$^3$}} & \multicolumn{3}{c}{\textbf{MMME}} \\ \cmidrule(r){2-4} \cmidrule(r){5-7} \cmidrule(r){8-10} \cmidrule(r){11-13} 
     & \textbf{F1-score} & \textbf{Precision} & \textbf{Recall} & \textbf{F1-score} & \textbf{Precision} & \textbf{Recall} & \textbf{F1-score} & \textbf{Precision} & \textbf{Recall} & \textbf{F1-score} & \textbf{Precision} & \textbf{Recall} \\ \hline
    MESNet (2021) \cite{wang2021mesnet} & 0.0490 & 0.0280 & 0.2300 & 0.0260 & 0.0130 & 0.1800 & - & - & - & 0.0511 & 0.0372 & 0.2963 \\
    ABPN (2022) \cite{leng2022abpn} & 0.2264 & 0.2727 & 0.1935 & 0.1590 & - & - & 0.3529 & 0.3750 & 0.3333 & 0.2861 & 0.3104 & 0.2547 \\
    SL-Swin (2023) \cite{he2023sl} & 0.0898 & 0.0689 &  0.1290 & 0.0879 & 0.0556 & 0.2105 & 0.1944 & 0.1944 & 0.1944 & 0.2152 & 0.2290 & 0.2694 \\
    DMA-LBP (2024) \cite{liu2024duration} & 0.0423 & 0.0230 & 0.2642 & 0.0096 & 0.0048 & 0.5263 & - & - & - & 0.1563 & 0.1350 & 0.2453 \\
    MSOF (2025) \cite{yang2025msof} & 0.3404 & 0.5000 & 0.2580 & 0.3928 & 0.3854 & 0.4005 & 0.3902 & 0.3478 & 0.4444 & 0.3747 & 0.4356 & 0.4122 \\ \hline
    \toprule[0.5pt]
    \end{tabular}%
    }
        \begin{tablenotes} 
            \fontsize{8pt}{8pt}\selectfont 
            \item[1] $^1$ The '-' in the table indicates that the information is not provided in the original paper.
        \end{tablenotes}
\end{table*}

\subsubsection{ME Spotting based on Multimodal Fusion}

Existing ME spotting algorithms primarily rely on facial visual information. Traditional approaches extract features such as facial landmark displacements, Local Binary Patterns from Three Orthogonal Planes (LBP-TOP), and optical flow vectors. More recent methods leverage deep learning architectures, including CNNs and Transformers, to model the spatiotemporal dynamics of MEs. However, due to the inherently low signal-to-noise ratio of ME signals and the susceptibility of manual annotations to bias, visual-based spotting algorithms often exhibit suboptimal performance and struggle to meet the requirements of real-world applications. Recent studies suggest that physiological responses hold potential value in ME spotting. For instance, Lu \textit{et al.} \cite{lu2022more} employed facial EMG to analyze the intensity characteristics of MEs and found that EMG can serve as a distinguishing indicator between micro- and macro-expressions. Similarly, Gupta \textit{et al.} \cite{gupta2018exploring} investigated the estimation of instantaneous heart rate from facial videos and demonstrated its utility in enhancing ME spotting performance. Given that human emotional changes are often accompanied not only by MEs but also by fluctuations in heart rate, the co-occurrence of MEs and instantaneous HRV has been widely recognized as a physiological manifestation of emotional arousal. This phenomenon has been extensively applied in lie detection technologies. 

\begin{figure}[b]
    \centering
    \includegraphics[width=0.95\linewidth]{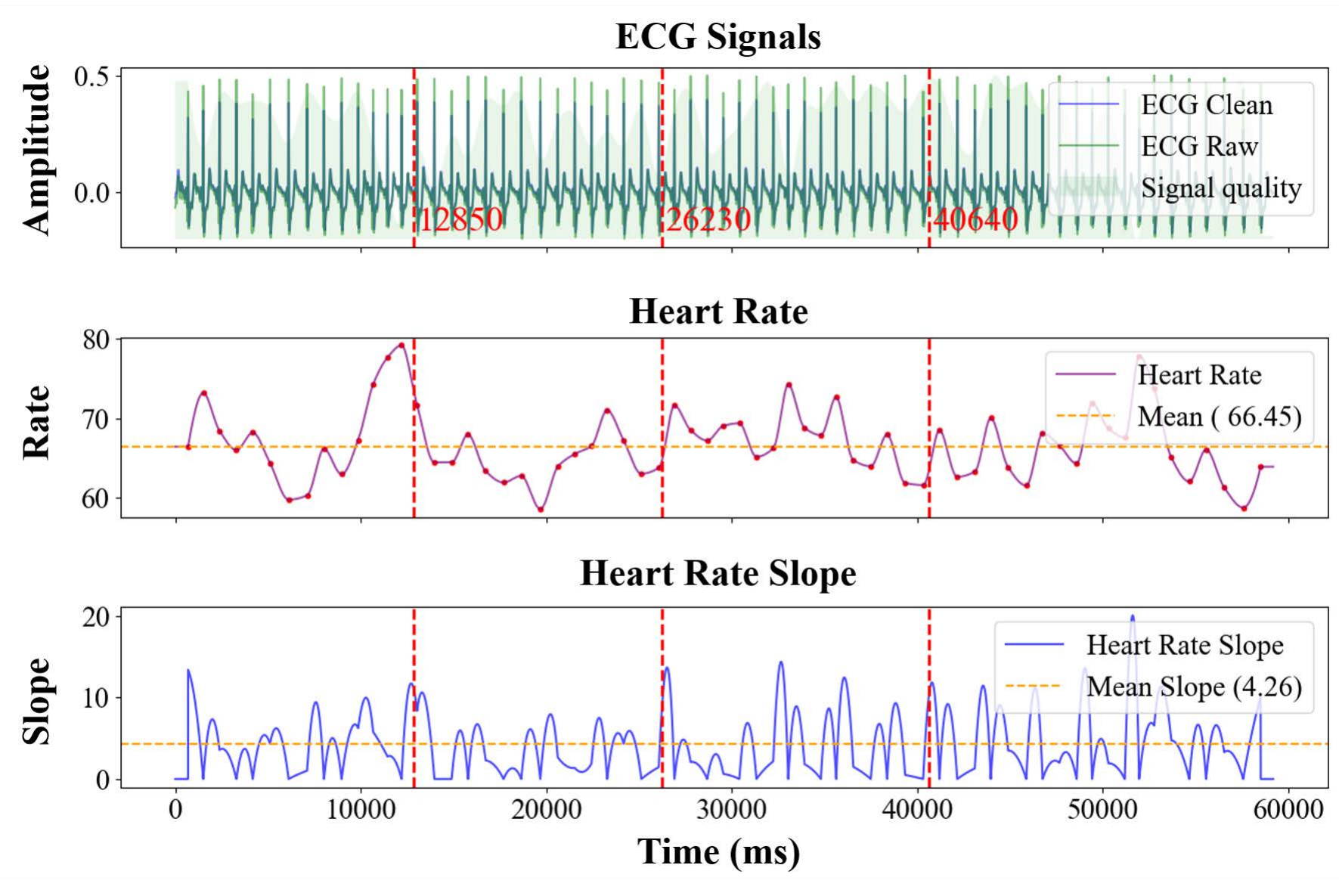}
    \caption{Relationship between MEs and heart rate in a representative experimental trial.}
    \label{fig:Heart Rate Slope.pdf}
\end{figure}

In Section \ref{subsection-concordance}, we analyzed the consistency between MEs and PS. The results reveal a statistically significant correlation (\textit{p} $<$ 0.05) between the occurrence of MEs and certain instantaneous heart rate variability features. This correlation is likely attributable to the fact that both MEs and heart rate responses are jointly regulated by emotional changes and the autonomic nervous system. Specifically, when an individual experiences intense emotional fluctuations, the sympathetic nervous system becomes activated, leading to measurable changes in heart rate. These findings provide a theoretical foundation for this study’s approach: investigating the use of ECG features to enhance the accuracy of ME spotting. Fig. \ref{fig:Heart Rate Slope.pdf} illustrates the relationship between ME occurrences and heart rate variability in a representative experimental trial: the upper panel displays the ECG signal waveform, the middle panel shows the temporal trend of heart rate, and the lower panel depicts the variation in heart rate slope. Red vertical lines indicate the precise timestamps of three ME apex frames in this trial, occurring at 12,850 ms, 26,230 ms, and 40,640 ms. Notably, while the absolute heart rate values at these ME events differ, each coincides with a region of elevated heart rate slope (HRS), reflecting instantaneous heart rate changes significantly above the trial’s mean. This observation suggests a potential link between ME apex frame and the HRS. To verify the generality of this phenomenon, a statistical analysis was conducted on all ME samples. The results revealed that, among 634 ME samples, 464 ME apex frames corresponded to HRS values exceeding the trial’s mean slope ($Mean_{HR\_slope}$), representing approximately 73.19\%. Additionally, 345 ME apex frames had HRS values surpassing 1.5 times the trial’s mean heart rate slope, accounting for approximately 54.41\%. These results further confirm a correlation between the occurrence of MEs and significant increases in HRS (i.e., rapid instantaneous changes in heart rate). 

\textbf{Candidate Clip Filter Module.} Building upon the aforementioned analysis, we propose a Candidate Clip Filter (CCF) module that integrates HRS features to authenticate detected ME clips: a candidate ME is classified as genuine if the HRS at its apex frame exceeds the trial’s mean slope value; conversely, if the slope is less than the mean value, it is classified as a spurious ME. The specific classification rule is detailed in Equation \ref{con: Genuine}:

    \begin{equation}
        \label{con: Genuine}
        \small
        Classify(apex_i) = 
        \begin{cases}
        Genuine, & \text{if } v_s > Mean_{HR\_slope} \\[1.2ex]
        Spurious, & \text{otherwise}
        \end{cases},
    \end{equation}
here, \textit{$apex_i$} denotes the apex frame of the $i$-th candidate ME, \textit{$v_s$} represents the slope of the heart rate curve at the apex frame, and $Mean_{HR\_slope}$ refers to the average slope of the heart rate curve within the same trial. 

\textbf{Experimental Results.} In this section, we evaluate the performance of various ME spotting algorithms on MMME dataset and further analyzes the optimization effects of the CCF module. As shown in Table \ref{tab: post-processing}, the original algorithms exhibit varying performance without CCF module, with MSOF achieving significantly higher F1-score (0.3747) and Precision (0.4356) compared to other methods. After integrating CCF, the precision of all algorithms improves (e.g., MSOF + CCF increases to 0.4790), while recall remains unchanged, confirming thatCCF enhances prediction reliability by effectively reducing FP. Concurrently, the F1-score shows universal improvement (e.g., SL-Swin + CCF rises from 0.2152 to 0.2780), demonstrating CCF module's efficacy in balancing precision and recall. The experimental results indicate that the CCF module is effective in enhancing the detection performance of moderately performing algorithms (e.g., ABPN, SL-Swin) while maintaining the superiority of high-performance algorithms (e.g., MSOF), thereby providing an efficient post-processing optimization solution for ME spotting. The findings of this section validate the effectiveness of the PS-assisted ME spotting algorithm, demonstrating the substantial potential of integrating visual information with ECG in ME spotting tasks.


\begin{table}[t]
    \centering
    \tiny
    \caption{Performance of different ME spotting algorithms combined with Candidate Clip Filter (CCF) module on the MMME dataset}
    \label{tab: post-processing}
    \resizebox{\columnwidth}{!}{%
    \begin{tabular}{lccc}
    \hline
    \toprule[0.4pt]
    \multicolumn{1}{c}{\textbf{Method}} & \textbf{F1-score} & \textbf{Precision} & \textbf{Recall} \\ \hline
    MESNet (2021) \cite{wang2021mesnet} & 0.0511 & 0.0372 & 0.2963 \\
    ABPN (2022) \cite{leng2022abpn} & 0.2861 & 0.3104 & 0.2547 \\
    SL-Swin (2023) \cite{he2023sl} & 0.2152 & 0.2290 & 0.2694 \\
    DMA-LBP (2024) \cite{liu2024duration} & 0.1563 & 0.1350 & 0.2453 \\
    MSOF (2025) \cite{yang2025msof} & 0.3747 & 0.4356 & 0.4122 \\ \hline
    MESNet + CCF & 0.1067 & 0.0914 & 0.2963 \\
    ABPN + CCF & 0.3646 & 0.3932 & 0.2547 \\
    SL-Swin + CCF & 0.2780 & 0.2949 & 0.2694 \\
    DMA-LBP + CCF & 0.1734 & 0.1602 & 0.2453 \\
    MSOF + CCF & 0.4023 & 0.4790 & 0.4122 \\ \hline
    \toprule[0.4pt]
    \end{tabular}%
    }
\end{table}

\section{Ethical Issues}

The research purpose and procedure were explained to each participant before the recording started, and the participants were well-aware that they can stop and quit the recording at anytime. One consent form was signed when the participant understood the contents and agreed to participate. Special questions were asked in the consent form concerning the data sharing issue, and the participants choose between two levels: 1) all recorded data could be shared and used for research analysis, and facial images and videos can be published or presented for academic purposes, e.g., in paper publications, presentations, webpages, or demos; 2) all recorded data could be shared and used for research analysis, but facial images and videos cannot be published or presented. 62 participants agreed on level-1, and the rest 13 participants agreed on level-2.

\section{Conclusion and Perspective}

\subsection{Conclusion}

This study addresses the limitations of micro-expression (ME) research, which has traditionally focused on a single visual modality while neglecting emotional information from other modalities. To overcome this, we developed the Multimodal ME Dataset (MMME), the most comprehensive and rigorously synchronized multimodal dataset to date. For the first time, this dataset enables the simultaneous collection of facial action signals (MEs), central nervous signals (EEG), and peripheral physiological signals (PS)---including PPG, RSP, SKT, EDA and ECG---thereby surpassing the constraints of existing ME corpora. The Continuous Monitoring and Real-Time Reminding (CMRR) paradigm proposed in this study aligns with the core characteristic of MEs as ``emotional leakage under controlled conditions.'' By minimizing interference, this paradigm achieves a balance between experimental control and ecological validity, significantly enhancing the efficiency of ME elicitation. The MMME dataset, meticulously annotated by multiple experts, comprises 634 MEs, 2,841 macro-expressions (MaEs), and 2,890 trials of multimodal PS. This high-quality dataset provides a robust foundation for exploring the neural mechanisms of MEs and conducting multimodal PS fusion analyses. Concordance analysis between MEs and PS reveals significant correlations between MEs and physiological features such as HRV, RSP, and EDA activity. These findings validate the existence of visual-physiological synergistic effects and offer a theoretical basis for multimodal fusion strategies. Extensive experiments confirm the effectiveness of the MMME dataset: in ME recognition tasks, the proposed multimodal fusion strategies leverage complementary information from different modalities, significantly improving recognition performance. In ME spotting tasks, integrating visual features with heart rate slope effectively eliminates spurious-ME segments, further enhancing detection accuracy. These outcomes advance ME research from single-modality visual analysis to a multimodal fusion paradigm, offering critical support for applications in intelligent human-computer interaction and mental health monitoring.

\subsection{Perspective}

At the data level, future work will focus on enriching the diversity of the dataset by incorporating participants from various ethnicities, cultural backgrounds, and age groups to enhance its generalizability and representativeness. Additionally, immersive interactive scenarios will be designed to construct datasets that are more reflective of real-world contexts. At the technical level, efforts will be directed toward advancing multimodal fusion algorithms from ``modality stacking'' to ``mechanism modeling.'' This will involve techniques such as semantic alignment of heterogeneous modal data, cross-modal attention mechanisms, and multimodal graph modeling to achieve temporal synchronization and semantic coordination across modalities, thereby advancing multimodal affective computing toward mechanism understanding and reasoning capabilities. At the theoretical level, the application of the ``multi-channel emotional expression theory" and the "emotional consistency theory" in the field of MEs will be further deepened. This will include exploring the neural mechanisms of MEs and uncovering the underlying principles of visual-physiological synergistic effects.

\section*{Acknowledgments}
This work was supported in part by the grants from the National Natural Science Foundation of China under Grant (No.62332019, No.62076250, No.62406338 and No.62204204), the National Key Research and Development Program of China (No.2023YFF1203900 and No.2023YFF1 203903), the Innovation Capability Support Program of Shaanxi (2024RS-CXTD-7), the Natural Science Basic Research Plan in Shaanxi Province of China (2023JC-XJ-07).

\printcredits

\bibliographystyle{unsrtnat}

\bibliography{cas-refs}

\end{document}